\documentclass{article}

\usepackage{microtype}
\usepackage{graphicx}
\usepackage{subcaption}
\usepackage{booktabs} 
\usepackage{multirow}
\usepackage{xcolor}

\usepackage{hyperref}

\usepackage[preprint]{icml2026}

\usepackage{amsmath}
\usepackage{amssymb}
\usepackage{mathtools}
\usepackage{amsthm}

\usepackage[capitalize, noabbrev]{cleveref}

\theoremstyle{plain}

\theoremstyle{definition}

\theoremstyle{remark}

\usepackage[textsize=tiny]{todonotes}

\usepackage{xcolor}
\usepackage{adjustbox}

\definecolor{color-algcomment}{rgb}{0,0.7059,0.1647}

\icmltitlerunning{}

\begin{document}
\captionsetup[subfigure]{labelformat=empty}
\captionsetup[subfigure]{font=scriptsize, skip=1pt}
\captionsetup[figure]{font=small, skip=2pt}

\twocolumn[
  \icmltitle{BinaryDemoire: Moir\'e-Aware Binarization for Image Demoiréing
    }



  \icmlsetsymbol{equal}{*}

  \begin{icmlauthorlist}
    \icmlauthor{Zheng Chen}{equal,sjtu}
    \icmlauthor{Zhi Yang}{equal,sjtu}
    \icmlauthor{Xiaoyang Liu}{sjtu}
    \icmlauthor{Weihang Zhang}{thu}
    \icmlauthor{Mengfan Wang}{vt}\\
    \icmlauthor{Yifan Fu}{thu}
    \icmlauthor{Linghe Kong}{sjtu}
    \icmlauthor{Yulun Zhang$^{\dagger}$}{sjtu}
  \end{icmlauthorlist}
  
  \icmlaffiliation{sjtu}{Shanghai Jiao Tong University}
  \icmlaffiliation{thu}{Tsinghua University}
  \icmlaffiliation{vt}{Virginia Tech}

  \icmlcorrespondingauthor{Yulun Zhang}{yulun100@gmail.com}

  \icmlkeywords{Binarization, Image Demoiréing}
  \vskip 0.3in
]



\printAffiliationsAndNotice{\icmlEqualContribution}

\begin{abstract}
Image demoir\'eing aims to remove structured moir\'e artifacts in recaptured imagery, where degradations are highly frequency-dependent and vary across scales and directions. While recent deep networks achieve high-quality restoration, their full-precision designs remain costly for deployment. Binarization offers an extreme compression regime by quantizing both activations and weights to 1-bit. Yet, it has been rarely studied for demoir\'eing and performs poorly when naively applied. In this work, we propose \textbf{BinaryDemoire}, a binarized demoir\'eing framework that explicitly accommodates the frequency structure of moir\'e degradations. First, we introduce a moir\'e-aware binary gate (MABG) that extracts lightweight frequency descriptors together with activation statistics. It predicts channel-wise gating coefficients to condition the aggregation of binary convolution responses. Second, we design a shuffle-grouped residual adapter (SGRA) that performs structured sparse shortcut alignment. It further integrates interleaved mixing to promote information exchange across different channel partitions.
Extensive experiments on four benchmarks demonstrate that the proposed BinaryDemoire surpasses current binarization methods. Code:~\url{https://github.com/zhengchen1999/BinaryDemoire}.
\end{abstract}

\begin{figure}
    \centering
    \includegraphics[width=1.0\linewidth]{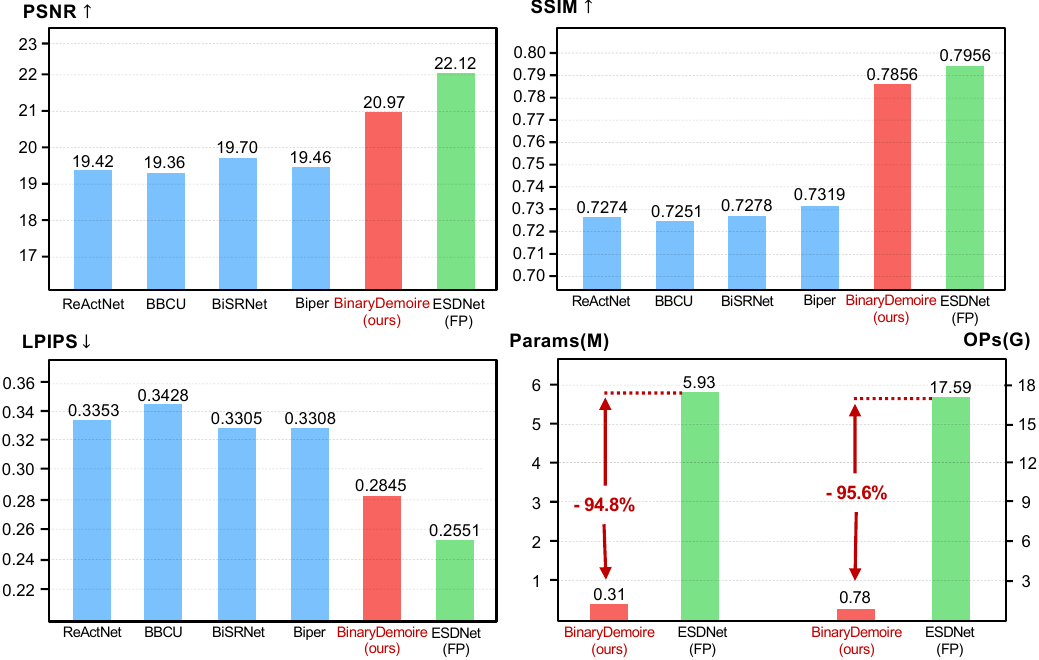}
    \caption{Performance and efficiency comparison between binarization methods and the full-precision baseline model (ESDNet~\cite{yu2022towards}). The results are evaluated on the UHDM dataset.}
    \label{fig:performance}
    \vspace{-8.mm}
\end{figure}

\begin{figure*}[!t]
\scriptsize
\centering
\begin{tabular}{cccccccc}
\hspace{-0.48cm}
\begin{adjustbox}{valign=t}
\begin{tabular}{c}
\includegraphics[width=0.2143\textwidth]{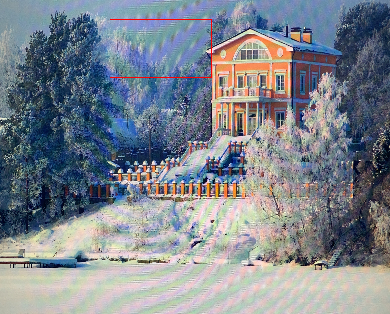}
\\
UHDM:0007
\end{tabular}
\end{adjustbox}
\hspace{-0.46cm}
\begin{adjustbox}{valign=t}
\begin{tabular}{cccccc}
\includegraphics[width=0.189\textwidth]{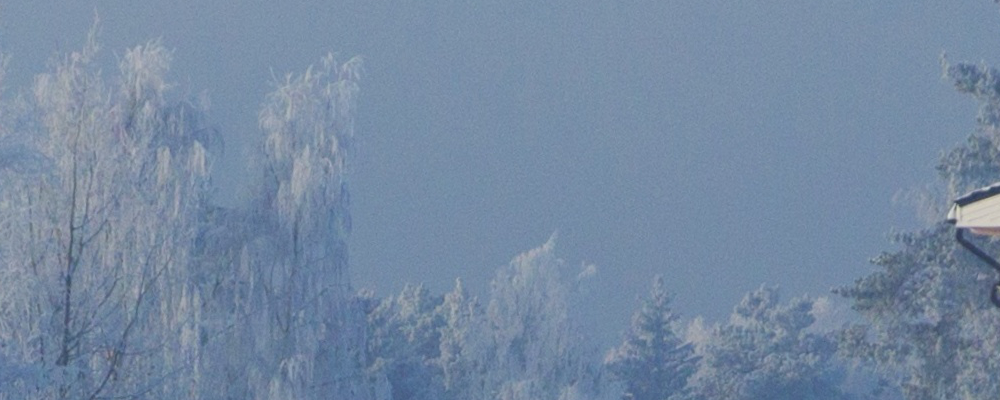} \hspace{-4.mm} &
\includegraphics[width=0.189\textwidth]{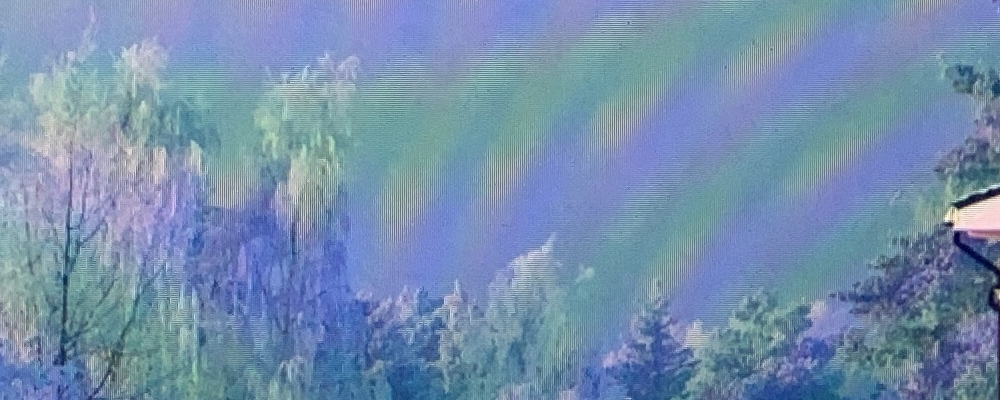} \hspace{-4.mm} &
\includegraphics[width=0.189\textwidth]{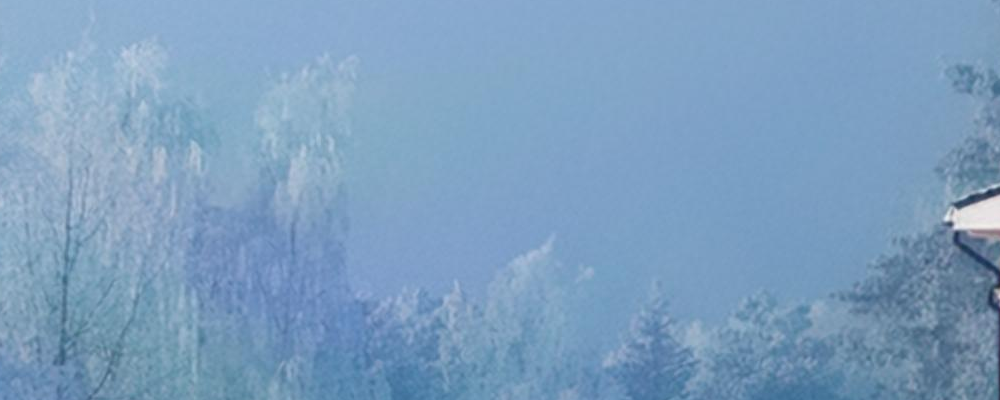} \hspace{-4.mm} &
\includegraphics[width=0.189\textwidth]{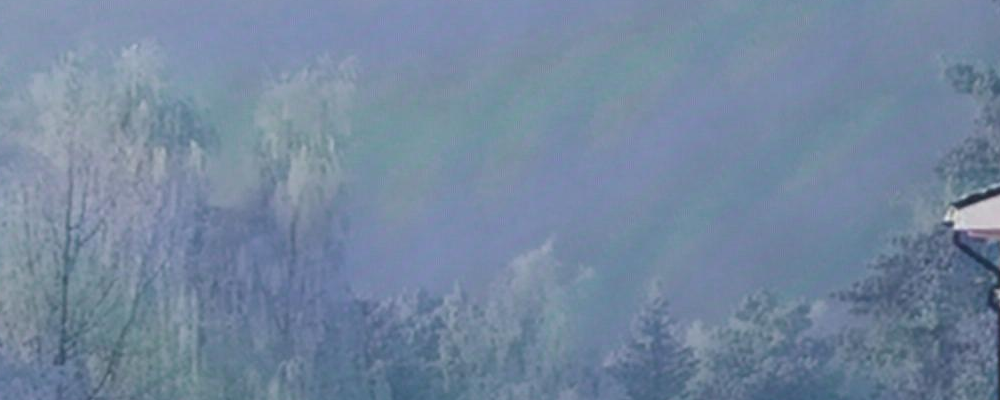} \hspace{-4.mm} &
\\ 
GT \hspace{-4.mm} &
INPUT \hspace{-4.mm} &
ESDNet (FP)~\cite{yu2022towards} \hspace{-4.mm} &
BBCU~\cite{xia2022basic} \hspace{-4.mm} &
\\
\includegraphics[width=0.189\textwidth]{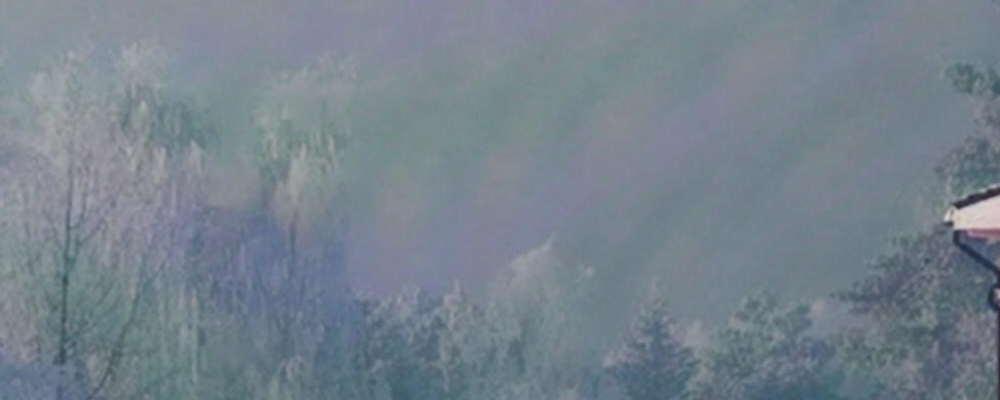} \hspace{-4.mm} &
\includegraphics[width=0.189\textwidth]{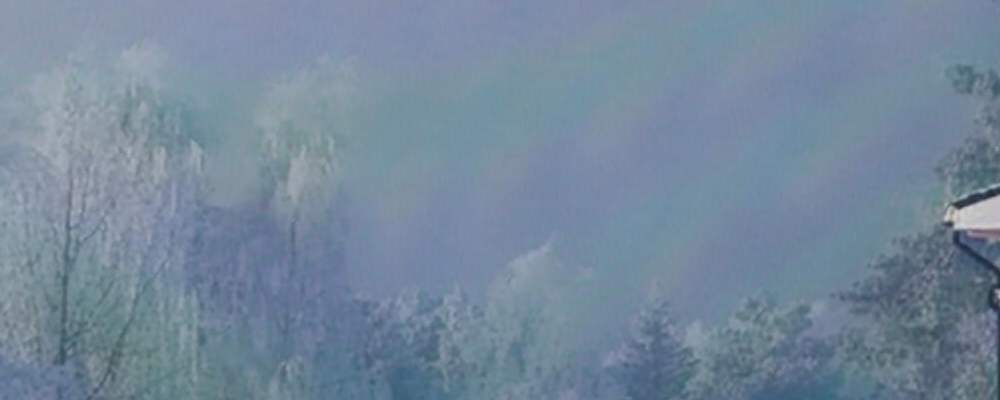} \hspace{-4.mm} &
\includegraphics[width=0.189\textwidth]{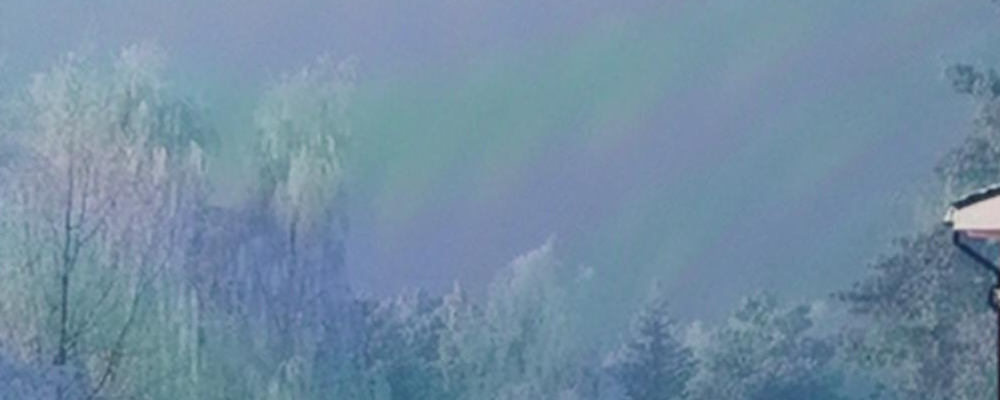} \hspace{-4.mm} &
\includegraphics[width=0.189\textwidth]{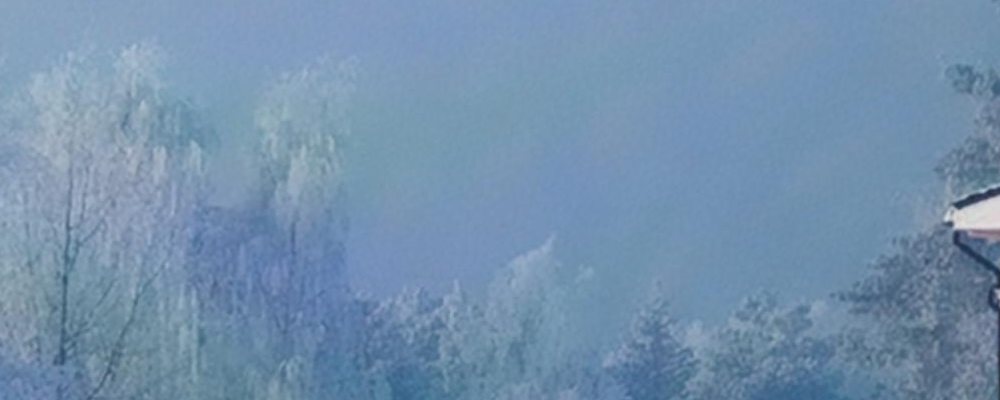} \hspace{-4.mm} &
\\ 
BiSRNet~\cite{cai2023binarized} \hspace{-4.mm} &
Biper~\cite{vargas2024biper} \hspace{-4.mm} &
BiMaCoSR~\cite{liu2025bimacosr} \hspace{-4.mm} &
BinaryDemoire (ours) \hspace{-4mm}
\\
\end{tabular}
\end{adjustbox}
\\

\end{tabular}
\caption{Visual comparison of binarization methods. Our proposed BinaryDemoire outperforms other methods with accurate results.}
\label{fig:performance-visual}
\end{figure*}

\vspace{-6.mm}
\section{Introduction}
\vspace{-1.mm}

Image demoir\'eing targets the removal of moir\'e artifacts frequently observed in recaptured images, particularly for screen-captured content. Existing analyses attribute these artifacts to structured aliasing caused by mismatched periodic sampling processes between the display and the camera imaging pipeline. It makes the moir\'e highly non-stationary and content-dependent~\cite{wang2023coarse}. Such degradations typically appear as directional, multi-scale high-frequency textures, substantially harming perceptual quality. This motivates demoir\'eing approaches that are both restoration-effective and deployment-efficient.

Recent advancements in deep neural networks have contributed to significant progress in image demoiréing. The demoir\'eing methods~\cite{he2020fhde2net,yu2022towards,zheng2020image} typically build upon multi-scale encoder-decoder backbones~\cite{ronneberger2015u}. They enhance cross-scale interaction via stage-wise aggregation~\cite{peng2024image}, or explicitly model moir\'e in the frequency domain through decomposition-based designs~\cite{liu2025freqformer}. Despite favorable restoration performance, these methods often rely on computationally heavy full-precision backbones and dense convolutions. This results in substantial storage and computational overhead. The high complexity makes them less suitable for practical deployment on edge platforms where memory footprint, latency, and energy consumption are tightly constrained.

Model compression and quantization provide a promising direction to bridge the gap between restoration quality and deployability. As an extreme form of quantization, binarization quantizes both activations and weights to 1-bit. It enables the replacement of expensive multiply-accumulate operations with bitwise XNOR-popcount computations, significantly reducing memory and computational cost. Driven by these efficiency benefits, binarization research has gradually expanded from classification to a broader range of low-level vision problems~\cite{xia2022basic,cai2023binarized,chen2024binarized,vargas2024biper,liu2025bimacosr}. 

However, binarizing demoir\'eing models remains largely under-explored. Applying existing binarization schemes directly to demoir\'eing backbones often yields suboptimal results. This gap mainly stems from \textbf{two} mismatches. 
\textbf{First}, many binarization methods~\cite{rastegari2016xnor,liu2020reactnet} adopt distribution-driven parameterization that is largely input-agnostic. However, moir\'e artifacts are highly structured, frequency-dominated degradations with strong directionality. Thus, uniform binarization incurs pronounced moir\'e-sensitive quantization errors. 
\textbf{Second}, binary networks~\cite{liu2018bi} rely on identity shortcuts to preserve a stable full-precision information path. Yet demoir\'eing architectures frequently change feature dimension in multi-scale processing, making plain residual connections inapplicable.
Although recent methods~\cite{liu2025bimacosr} introduce low-rank full-precision branches to maintain such information flow, their overhead can still be non-negligible. This issue becomes more pronounced when the overall channel width is relatively small, which is common in lightweight restoration backbones.

\vspace{-0.5mm}
To address the above challenges, we propose \textbf{BinaryDemoire}, an effective binarized demoir\'eing framework. BinaryDemoire tackles the two mismatches from complementary perspectives. 
For the \textbf{first} issue, we introduce the moir\'e-aware binary gate (MABG). The module extracts lightweight frequency descriptors together with activation statistics from full-precision features. It then predicts channel-wise gates to condition the aggregation of binary convolution responses.
For the \textbf{second} issue, we design the shuffle-grouped residual adapter (SGRA) as a lightweight shortcut alignment module.
It performs structured sparse 1$\times$1 projection under channel partitioning and integrates an interleaved permutation to promote cross-partition information exchange.
As a result, SGRA provides a general and efficient solution for residual alignment under frequent dimensional mismatches, avoiding costly auxiliary branches.

\vspace{-0.5mm}
We conduct massive experiments on four demoiréing benchmark datasets. As shown in Figs.~\ref{fig:performance} and \ref{fig:performance-visual}, our method significantly outperforms existing binarized approaches. In comparison with the full-precision model~\cite{yu2022towards}, it achieves comparable performance while reducing more than 90\% of the parameters and computational cost. These results demonstrate the advantages of BinaryDemoire.

In general, our contributions can be outlined as follows:
\begin{itemize}
\vspace{-1.mm}
\item We propose BinaryDemoire, a binarized framework for image demoiréing. To the best of our knowledge, this is the first work to introduce binarization into demoiréing.

\vspace{-1.mm}
\item We design the moiré-aware binary gate (MABG) to adapt to diverse inputs, and the shuffle-grouped residual adapter (SGRA) for easy information propagation.

\vspace{-1.mm}
\item Extensive comparisons with existing binarized methods across diverse benchmarks demonstrate the superior performance of BinaryDemoire.
\end{itemize}

\section{Related Work}

\begin{figure*}
    \centering
    \includegraphics[width=1.\linewidth]{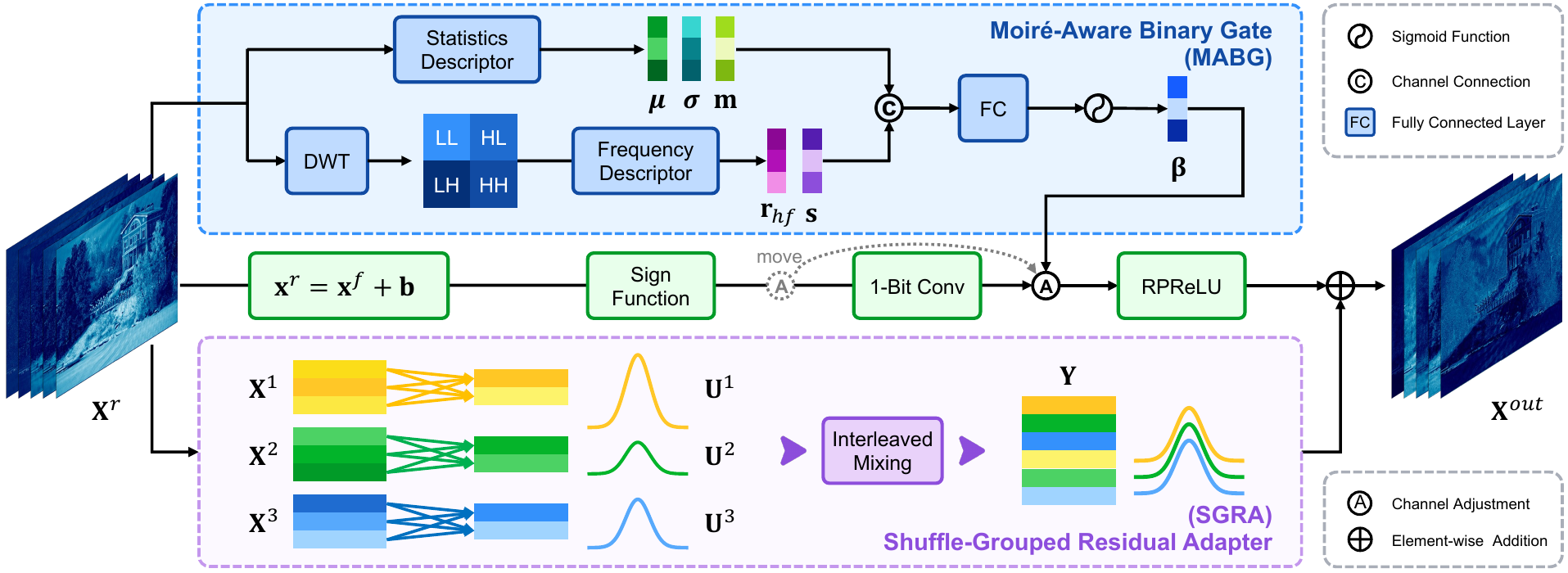}
    \vspace{1.mm}
    \caption{Overview of the BinaryDemoire framework. The BinaryDemoire contains two modules. The first is moiré-aware binary gate (MABG), which adapts binary activation through frequency and statistical descriptors to effectively modulate the binary gate values. The second module is the shuffle-grouped residual adapter (SGRA), which provides lightweight residual alignment across channel and spatial dimensions, enabling effective feature exchange and preserving critical information during the demoiréing process.}
    \label{fig:placeholder2}
\end{figure*}

\textbf{Image Demoiréing.}
Image demoiréing aims to remove moiré patterns from captured images. Early demoiréing methods primarily relied on handcrafted priors and traditional signal processing techniques such as smooth filtering~\cite{siddiqui2009hardware}, spectral model~\cite{sidorov2002suppression}, and sparse matrix decomposition~\cite{liu2015moire}. These methods typically target specific moiré textures and generalize poorly to diverse real-world patterns. With recent development in deep learning, end-to-end neural methods~\cite{cheng2019multi,he2019mop,yu2022towards,xu2023image,liu2025freqformer} have been widely adopted for image demoiréing, delivering improved performance. FHDe²Net~\cite{he2020fhde2net} targets practical high-resolution demoiréing via a cascaded design and frequency-guided content separation.
To improve robustness in ultra-high-definition settings, ESDNet~\cite{yu2022towards} emphasizes scale robustness and efficiency on 4K data.
Meanwhile, Freqformer~\cite{liu2025freqformer} builds on efficient frequency decomposition to better disentangle high-frequency texture corruption.
However, many state-of-the-art demoiréing models remain computationally heavy, making efficient deployment still challenging.

\textbf{Binarization.}
Binarization typically compresses the network into 1-bit weights and activations, where parameters are represented as ±1, and most multiplications can be replaced by bitwise operations. Binarization research was initially in high-level tasks~\cite{hubara2016binarized,rastegari2016xnor,liu2018bi,liu2020reactnet,vargas2024biper}, which mainly improve the accuracy via improved binary operators and training strategies. Recently, researchers have started to apply binarization to low-level vision tasks~\cite{jiang2021training,zhang2024flexible,wei2025scales}. BBCU~\cite{xia2022basic} revisits key components in binary convolutions for restoration and proposes a simple binary convolution unit that is effective across typical IR settings. BiSRNet~\cite{cai2023binarized} further extends binarization to spectral compressive imaging by introducing binarization-aware spectral redistribution. BI-DiffSR~\cite{chen2024binarized} designs binarization-friendly UNet modules and timestep-aware activation handling. BiMaCoSR~\cite{liu2025bimacosr} combines binarization with one-step distillation and lightweight full-precision auxiliary branches to prevent collapse. Nevertheless, binarized low-level vision models often still suffer from noticeable performance degradation. As far as we are aware, binarization has not yet been systematically investigated for image demoiréing, motivating us to design a binarized demoiréing model tailored to moiré artifacts.

\section{Methodology}
In this section, we introduce our binary demoiréing framework built on an ESDNet backbone~\cite{yu2022towards} for efficient deployment. We first present the overall framework, explaining how we selectively binarize the ESDNet-based pipeline to improve restoration quality and balance computational efficiency. Then, we detail two key designs: the moiré-aware binary gate (MABG), which exploits frequency and statistics descriptors to adaptively modulate binarized activations, and the shuffle-grouped residual adapter (SGRA), which provides lightweight residual alignment under dimensional mismatches.

\subsection{Overall Framework}
\label{sec:overall}

We propose a deployment-friendly binary image demoir\'eing framework built upon ESDNet~\cite{yu2022towards}, a multi-scale encoder-decoder backbone. We focus on binarizing its full-precision convolutions while maintaining restoration fidelity. Specifically, except for the first convolution (which stabilizes the input feature distribution) and the final convolution (which preserves output fidelity), all intermediate convolutional layers are binarized using 1-bit weights and activations. As a result, the main computation is carried out using binarized weights and activations.

Given a full-precision feature tensor $\mathbf{X}^{f}$$\in$$\mathbb{R}^{B\times C\times H\times W}$ and weight $\mathbf{W}^{f}$$\in$$\mathbb{R}^{C_{\text{out}}\times C_{\text{in}}\times K\times K}$, our binarized block (Fig.~\ref{fig:placeholder2}) consists of a binary main branch and a shortcut branch. 

\textbf{Binarization.} We first apply a learnable channel-wise threshold $\mathbf{t}$$\in$$\mathbb{R}^{1\times C\times 1\times 1}$~\cite{liu2020reactnet} to $\mathbf{X}^{f}$ and then binarize them with sign function.
\begin{equation}
\mathbf{X}^{b}=\operatorname{Sign}\!\left(\mathbf{X}^{f}+\mathbf{t}\right), \mathbf{W}^b = \operatorname{Sign}(\mathbf{W}^f),
\label{eq:act_bin_overall}
\end{equation}

Let $\mathbf{X}^{b}$$\in$$\{\pm1\}^{C_{\mathrm{in}}\times H\times W}$ represent the input feature map and
$\mathbf{W}^b$$\in$$\{\pm1\}^{C_{\mathrm{out}}\times C_{\mathrm{in}}\times K\times K}$ the binarized convolutional kernels. We introduce an input-channel gate vector $\boldsymbol{\beta}$$\in$$\mathbb{R}^{C_{\mathrm{in}}}$  which is predicted from the full-precision input by MABG module and a per-output-channel scaling factor
$\boldsymbol{\alpha}$$\in$$\mathbb{R}^{C_{\mathrm{out}}}$ ($\boldsymbol{\alpha}=\|\mathbf{W}^f\|_1/(C_{\mathrm{in}}K^2)$, following the XNOR-Net~\cite{rastegari2016xnor}).
Thus, the channel adjustment convolution output can be written as follows:
\begin{equation}
\mathbf{X}^b_{{out}} = (\boldsymbol{\beta} \mathbf{X}^b) * (\boldsymbol{\alpha} \mathbf{W}^b) = \boldsymbol{\alpha} \sum_{c=1}^{C_{\mathrm{in}}} \beta_c \, (\mathbf{X}^b_c \ast \mathbf{W}^b_{c}), 
\label{eq:gcbc}
\end{equation}
where $\ast$ denotes 2D convolution. The gating coefficients $\beta_c$ modulate only the channel-wise aggregation of convolution, while the inner products $\mathbf{X}^b_c\ast \mathbf{W}^b_c$ retain the binary structure. They can be computed via bitwise operations when both activations and weights are 1-bit.

The binary branch performs convolution followed by RPReLU, and the output is obtained via residual learning:
\begin{equation}
\mathbf{X}^{out} = \operatorname{RPReLU}\!(\mathbf{X}^b_{{out}}) + \operatorname{SGRA}\!\left(\mathbf{X}^{f}\right),
\label{eq:block_overall}
\end{equation}
$\operatorname{SGRA}(\cdot)$ is an identity mapping when the shortcut shape matches the main branch, otherwise it applies the proposed shuffle-grouped residual adapter for lightweight projection under channel mismatches (Sec.~\ref{sec:sgra}).

\textbf{Training.}
    Since $\operatorname{Sign}(\cdot)$ is not differentiable, we adopt straight-through estimators (STE)~\cite{bengio2013estimating} for stable end-to-end training. For activation binarization, we use a smooth surrogate $\tanh(\beta x)$ in the backward pass, where $\beta$ is learnable~\cite{cai2023binarized}.
    
    For weight binarization, we employ a clipped STE to propagate gradients through $\mathbf{W}^{b}$, following common practice in binary networks~\cite{courbariaux2015binaryconnect}.

\vspace{-1.mm}
\subsection{Moir\'e-Aware Binary Gate (MABG)}
\label{sec:mabg}
\vspace{-1.mm}

\textbf{Motivation.}
Most existing binarization schemes rely on shared, distribution-driven parameterization~\cite{liu2020reactnet,zhang2022dynamic,qin2023distribution,zhang2024binarized} and largely ignore the frequency structure of task-specific degradations. While this paradigm is often adequate for classification, it is less suitable for image demoiréing. Moiré artifacts typically appear as high-frequency, texture-like patterns with pronounced directionality and substantial variation across scales~\cite{he2020fhde2net,campos2024moirewidgets}. Effective demoiréing,  therefore, requires preserving frequency-sensitive information to disentangle moiré components from genuine image details. 

\textbf{Design.} Motivated by this mismatch, we propose a wavelet-based moiré-aware binary gate (MABG) that extracts frequency and statistics descriptors and uses them to adjust the binary convolution in a channel-wise manner. The key idea is to explicitly leverage wavelet-derived frequency information to adapt the effective amplitude of binarized activations in a channel-wise manner. Specifically, MABG predicts a gate value $\boldsymbol{\beta}_{b, c}$ from both (i) activation statistics and (ii) wavelet frequency descriptors, and then modulates the binarized convolution by $\boldsymbol{\beta}_{b, c}$. In this way, channels that are more sensitive to moir\'e frequency patterns can be assigned a more suitable binarization strength, alleviating the severe information loss caused by uniform binarization. We present the detailed procedure of MABG in Algorithm~\ref{alg:mabg}. Worth noting is that, to balance computational cost and effectiveness, MABG is applied selectively to key layers, with the specific allocation strategy provided in the supplementary material.

\begin{algorithm}[tb]
  \caption{Moir\'e-aware Binary Gate (MABG)}
  \label{alg:mabg}
  \begin{algorithmic}
    \STATE {{Input:}} full-precision feature $\mathbf{X}^f\in\mathbb{R}^{B\times C_{\mathrm{in}}\times H\times W}$
    \STATE {{Output:}} channel-wise gates $\mathbf{k}\in\mathbb{R}^{B\times C_{\mathrm{in}}}$

    \vspace{6pt}
    \STATE \textit{\textcolor{color-algcomment}{\# 1) Frequency descriptor}}
    \STATE $(\mathbf{X}_{HH}, \mathbf{X}_{HL}, \mathbf{X}_{LH}, \mathbf{X}_{LL}) \leftarrow \operatorname{DWT}(\mathbf{X}^f)$
    \FOR{$s \in \{HH, HL, LH, LL\}$}
      \STATE \hspace{1em}$\mathbf{E}_s \leftarrow \operatorname{MeanAbs}(\mathbf{X}_s)$
    \ENDFOR
    \STATE $\mathbf{E} \leftarrow \{\mathbf{E}_{HH}, \mathbf{E}_{HL}, \mathbf{E}_{LH}, \mathbf{E}_{LL}\}$
    \STATE $\mathbf{r}_{hf}, \mathbf{s} \leftarrow \operatorname{Frequency\text{-}Descriptor}(\mathbf{E})$
    
    \vspace{6pt}
    \STATE \textit{\textcolor{color-algcomment}{\# 2) Statistics descriptor}}
    \STATE $\boldsymbol{\mu} \leftarrow \operatorname{Mean}(\mathbf{X}^f)$
    \STATE $\boldsymbol{\sigma} \leftarrow \operatorname{Std}(\mathbf{X}^f)$
    \STATE $\mathbf{m} \leftarrow \operatorname{MeanAbs}(\mathbf{X}^f)$

    \vspace{6pt}
    \STATE \textit{\textcolor{color-algcomment}{\# 3) Gate prediction from frequency \& statistics}}
    \STATE $\mathbf{c} \leftarrow
    \operatorname{Concat}(\boldsymbol{\mu}, \boldsymbol{\sigma}, \mathbf{m}, \mathbf{r}_{hf}, \mathbf{s})$
    \STATE $\boldsymbol{\beta} \leftarrow
    \operatorname{Sigmoid}\!\big(\operatorname{FC}(\mathbf{c})\big)$

    \STATE return $\boldsymbol{\beta}$
  \end{algorithmic}
\end{algorithm}
\textbf{Frequency Descriptor.}
Given a full-precision activation tensor $\textbf{X}$$\in$$\mathbb{R}^{B\times C\times H\times W}$, we use a single-level discrete wavelet transform (DWT)~\cite{yeh2024multibranch,jiang2023fabnet} to obtain four sub-bands $\{HH, HL, LH, LL\}$, where $\mathbf{X}_i(i$$\in$$\{HH, HL, LH, LL\})$$\in$$\mathbb{R}^{C \times h \times w}$ denotes each frequency component, and $h=H/2, w=W/2$.

For each sub-band $s\in\{HH, HL, LH, LL\}$, we compute the mean absolute response as the energy proxy:
\begin{equation}
\mathbf{E}_{s}=\frac{1}{HW}\sum_{i=1}^{H}\sum_{j=1}^{W}\left|\mathbf{X}_{s}(i, j)\right|,
\label{eq:mabg_energy}
\end{equation}
Based on these energies, we define two frequency descriptors. The first is the high-frequency ratio $\mathbf{r}_{hf}$, measuring the proportion of high-frequency energy:
\begin{equation}
\mathbf{r}_{hf}=\frac{\mathbf{E}_{LH}+\mathbf{E}_{HL}+\mathbf{E}_{HH}}
{\mathbf{E}_{LL}+\mathbf{E}_{LH}+\mathbf{E}_{HL}+\mathbf{E}_{HH}},
\label{eq:mabg_rhf}
\end{equation}
The second descriptor captures the dominant orientation tendency of moir\'e-like patterns by comparing horizontal with vertical high-frequency bands:
\begin{equation}
\mathbf{s}=\frac{\max(\mathbf{E}_{LH}, \mathbf{E}_{HL})}{\mathbf{E}_{LH}+\mathbf{E}_{HL}},
\label{eq:mabg_orient}
\end{equation}
$\mathbf{r}_{hf}$ reflects how strongly a channel responds to high-frequency structures, while $\mathbf{s}$ indicates whether the response is dominated by a specific orientation.

\textbf{Statistics Descriptor.}
Besides frequency descriptor, we also summarize the activation distribution by per-sample, per-channel statistics: mean $\boldsymbol{\mu}_{b, c}$, standard deviation $\boldsymbol\sigma_{b, c}$, and mean absolute value $\boldsymbol{m}_{b, c}$ through a statistics descriptor, provide valuable insights into the distribution of feature activations, allowing the network to modulate binary activations more effectively~\cite{zhang2024binarized}.

\begin{figure}
    \centering
    \includegraphics[width=1.0\linewidth]{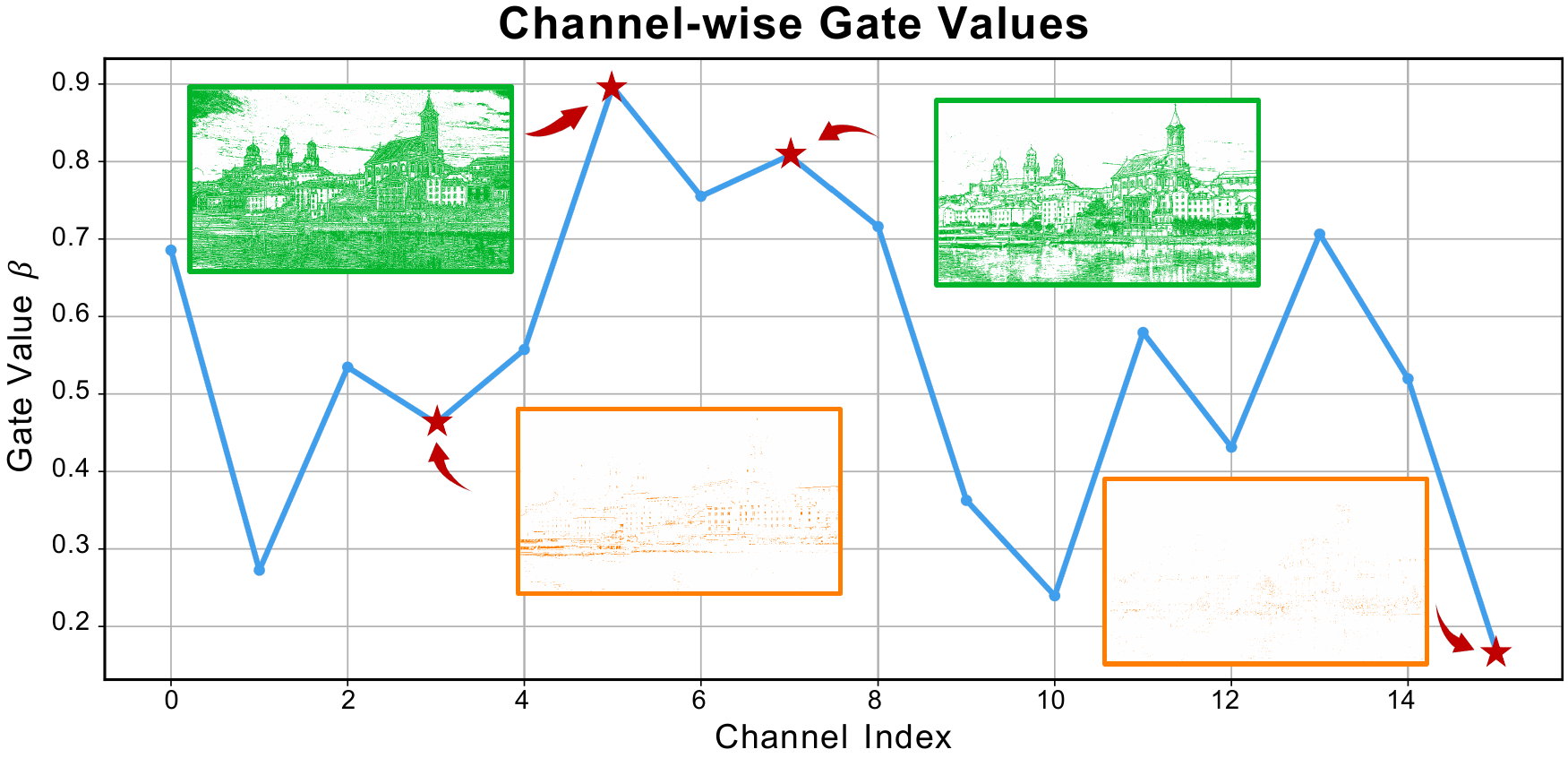}
    \caption{Channel-wise gate values visualization. We visualize the gate values corresponding to different channels, along with their associated binarized feature maps. For channels containing richer high-frequency details, the module learns larger gating responses to increase their contribution. This adaptive strategy highlights informative features and filters out less useful information.}
    \label{fig:gate}
    \vspace{-4.mm}
\end{figure}

\textbf{Gate Prediction From Frequency \& Statistics.}
After concatenating statistics descriptors with the frequency descriptors to form a compact vector, a shared fully connected layer followed by a sigmoid then maps each condition vector to a channel-wise gate~\cite{hu2018squeeze}. This design is lightweight due to parameter sharing, yet yields input-adaptive gates for modulating the binary convolution.
\begin{equation}
\boldsymbol{\beta}=\sigma\!\Big(\mathrm{FC}\big([\boldsymbol\mu, \, \boldsymbol\sigma, \, \boldsymbol{m}, \, \mathbf{r}_{hf}, \, \mathbf{s}]^\top\big)\Big).
\label{eq:mabg_gate}
\end{equation}

Here, $\mathrm{FC}(\cdot)$ denotes a lightweight shared fully connected layer applied to each per-channel condition vector, and $\sigma(\cdot)$ denotes the sigmoid function.

Figure.~\ref{fig:gate} visualizes the learned channel-wise gate values in our BinaryDemoire framework. The plot shows how the gate values vary across different channels, where the highlighted regions represent channels with higher or lower gate values. Channels with more prominent features (shown in green) are assigned higher gate values, while those with less significant features (shown in orange) receive lower gate values. This selective modulation helps mitigate the information loss caused by binary activation and improves the network sensitivity to frequency information, particularly in the context of image demoiréing.

\subsection{Shuffle-grouped Residual Adapter (SGRA)}
\label{sec:sgra}

\textbf{Motivation.}
Residual connections are particularly important for binary networks~\cite{liu2018bi,bethge2019binarydensenet,qin2020binary}, as the identity shortcut provides a stable information path and mitigates the severe representational loss introduced by 1-bit operations. However, in practical demoir\'eing backbones, channel dimensions and spatial resolutions change frequently due to downsampling, upsampling, or stage-wise feature aggregation. In these cases, a plain identity shortcut is no longer applicable, and a naive full-precision $1\times1$ projection introduces non-negligible parameters and computation. Although several methods have been suggested to address this issue, they often deliver suboptimal performance or exhibit limited generalization ability~\cite{chen2024binarized,liu2025bimacosr}. Therefore, we need a lightweight residual alignment mechanism that preserves the benefits of residual learning under frequent dimensional mismatches.

\textbf{Design.} To address the above issue, we introduce the shuffle-grouped residual adapter (SGRA) as a lightweight shortcut adapter for residual alignment under frequent dimension transitions. SGRA uses a structured sparse channel mapping to substantially reduce parameters and computation, while explicitly maintaining cross-partition information flow through an integrated interleaving scheme. Concretely, our design consists of two complementary components: (i) a partition-wise projection that performs efficient dimension matching within channel subspaces, and (ii) an interleaved mixing strategy that reorganizes channels to promote inter-partition feature exchange before fusion. 

\textbf{Partition-wise Projection.}
Let $\mathbf{X}$$\in$$\mathbb{R}^{B\times C_{\mathrm{in}}\times H\times W}$ be the shortcut input and
$\mathbf{U}$$\in$$\mathbb{R}^{B\times C_{\mathrm{out}}\times H'\times W'}$ the projected feature before mixing.
We set the partition number to the greatest common divisor of the input and output channels, 
$g=\gcd(C_{\mathrm{in}}, C_{\mathrm{out}})$, so that both $C_{\mathrm{in}}$ and $C_{\mathrm{out}}$ can be evenly
partitioned into $g$ disjoint channel subspaces.

Denote $C_{\mathrm{in}}$$=$$g\cdot c_{\mathrm{in}}$ and $C_{\mathrm{out}}$$=$$g\cdot c_{\mathrm{out}}$.
We perform a structured sparse projection independently within each partition:
\begin{equation}
\mathbf{U}^{(k)}(:, i, j)=\mathbf{W}^{(k)}\, \mathbf{X}^{(k)}(:, si, sj),
\label{eq:sgra_part_proj}
\end{equation}

$\mathbf{W}^{(k)}$$\in$$\mathbb{R}^{c_{\mathrm{out}}\times c_{\mathrm{in}}}, \;k=1, \ldots, g, $
where $\mathbf{X}^{(k)}$ and $\mathbf{U}^{(k)}$ denote the $k$-th channel partition of $\mathbf{X}$ and $\mathbf{U}$, 
respectively, and $s$ is the stride used to match spatial resolution when needed.
This partition-wise projection serves as a compact residual alignment operator, producing shortcut features that are
shape-compatible with the main branch by jointly adjusting channel width and spatial resolution.

\textbf{Interleaved Mixing.}
A purely partition-wise projection may lead to isolated representations, which limit information flow across channel subspaces. Inspired by the work~\cite{zhang2018shufflenet}, we apply an interleaved channel reordering to the projected feature. Let $m=C_{\mathrm{out}}/g$ and index channels by a partition index $k\in\{0, \ldots, g-1\}$ and an intra-partition
index $t\in\{0, \ldots, m-1\}$. The interleaving is defined as
\begin{equation}
\mathbf{Y}_{:, \, t\cdot g+k, \, i, \, j}=\mathbf{U}_{:, \, k\cdot m+t, \, i, \, j}.
\label{eq:interleave}
\end{equation}
which alternates channels from different partitions in the output layout, thereby promoting inter-partition feature exchange before fusion to the binary convolution branch.

Overall, SGRA provides a unified shortcut adaptation paradigm for binarized networks under shape transitions. By combining partition-wise structured projection with interleaved mixing, it produces shape-compatible residual features, retaining full-precision information. It does not rely on costly dense projections or case-specific designs. As a result, SGRA can be seamlessly inserted wherever dimensional mismatches occur, enabling consistent residual learning across the entire multi-scale demoir\'eing backbone.

\section{Experiments}

\subsection{Experimental Settings}
\label{sec:setting}
\vspace{-1.mm}

\textbf{Datasets and Metrics.} We perform experiments on four benchmark datasets: UHDM~\cite{yu2022towards}, FHDMi~\cite{he2020fhde2net}, LCDMoire~\cite{yuan2019aim}, and TIP2018~\cite{sun2018moire}. We evaluate the models using PSNR, SSIM~\cite{wang2004image}, and PIPS~\cite{zhang2018unreasonable} metrics for quantitative assessment. For computational cost, the number of parameters(\textbf{Params}) is computed as
\(
\text{Params} = \text{Params}^b + \text{Params}^f, 
\)
and the number of operations(\textbf{OPs}) is computed as
\(
\text{OPs} = \text{OPs}^b + \text{OPs}^f, 
\)
where $\text{Params}^b = {\text{Params}^f}/{32}$ and $\text{OPs}^b = {\text{OPs}^f}/{64}$, with the superscripts $f$ and $b$ representing full-precision and binarized modules. This evaluation setup follows previous works on binarized models~\cite{qin2023bibench}.

\textbf{Implementation Details.}
We adopt ESDNet~\cite{yu2022towards} as the backbone of our method. All convolution layers, except for the first and last layers, are binarized to 1-bit. The groups number of SGRA is set as \(g = \gcd(C_{\mathrm{in}}, C_{\mathrm{out}})\), where \(C_{\mathrm{in}}\) and \(C_{\mathrm{out}}\) represent the input and output channel dimensions of the respective convolution layer. 

\textbf{Training Settings.} For all four datasets (UHDM, FHDMi, LCDMoir\'e, and TIP2018), we follow the training procedure of ESDNet~\cite{yu2022towards} as closely as possible to ensure reproducibility and fair comparison. 
In particular, we randomly crop $768\times768$ patches from training images, set the batch size to 2, and train the models for 150 epochs. 
We optimize the network using Adam with $(\beta_1, \beta_2)=(0.9, 0.999)$, and set the initial learning rate to $2\times10^{-4}$, which is scheduled by a cyclic cosine annealing strategy~\cite{loshchilov2016sgdr}. 
All competing methods and our approach are trained under the same data preprocessing and optimization settings on each dataset unless stated otherwise.

\begin{table*}[t]
\centering
\scriptsize
\setlength{\tabcolsep}{7.0pt}
\renewcommand{\arraystretch}{1.12}
\caption{Ablation studies of BinaryDemoire. We report Params (M), OPs (G), and restoration quality (PSNR/SSIM/LPIPS).}
\label{tab:ablation_all}

\subfloat[{Break-down ablation.}\label{tab:ablation_break}]{
\setlength{\tabcolsep}{2.5mm}
\begin{tabular}{l | cc ccc}
\toprule
\textbf{Method} & {Params(M)} & {Ops(G)} & {PSNR} & {SSIM} & {LPIPS} \\
\midrule
Base        & 0.27 & 0.66 & 19.36 & 0.7407 & 0.3422 \\
+MABG       & 0.27 & 0.66 & 20.18 & 0.7562 & 0.3193 \\
+SGRA       & 0.31 & 0.78 & 20.68 & 0.7820 & 0.2912 \\
+MABG\&SGRA & 0.31 & 0.78 & \textbf{20.97} & \textbf{0.7856} &\textbf{0.2845} \\
\bottomrule
\end{tabular}
}
\hfill
\subfloat[{Ablation on MABG.}\label{tab:ablation_mabg}]{
\setlength{\tabcolsep}{2.5mm}
\begin{tabular}{l | cc ccc}
\toprule
\textbf{Method} & {Params(M)} & {Ops(G)} & {PSNR} & {SSIM} & {LPIPS} \\
\midrule
Freq only   & 0.31 & 0.78 & 20.72 & 0.7820 & 0.2896 \\
Stats only  & 0.31 & 0.78 & 20.72 & 0.7841 & 0.2904 \\
Learnable   & 0.32 & 0.78 & 20.75 & 0.7821 & 0.2990 \\
Freq\&Stats  & 0.31 & 0.78 & \textbf{20.97} & \textbf{0.7856} & \textbf{0.2845} \\
\bottomrule
\end{tabular}
}

\vspace{4pt}

\subfloat[{Ablation on SGRA.}\label{tab:ablation_sgra}]{
\setlength{\tabcolsep}{2.5mm}
\begin{tabular}{l | cc ccc}
\toprule
\textbf{Method} & {Params(M)} & {Ops(G)} & {PSNR} & {SSIM} & {LPIPS} \\
\midrule
w/o SGRA        & 0.27 & 0.66 & 20.18 & 0.7562 & 0.3193 \\
Low-rank ($r{=}2$) & 0.34 & 0.85 & 20.87 & 0.7820 & 0.2886 \\
w/o Mix          & 0.31 & 0.78 & 20.84 & 0.7841 & 0.2884 \\
SGRA               & 0.31 & 0.78 & \textbf{20.97} & \textbf{0.7856} & \textbf{0.2845} \\
\bottomrule
\end{tabular}
}
\hfill
\subfloat[{Ablation of group number.}\label{tab:ablation_gcd}]{
\setlength{\tabcolsep}{2.8mm}
\begin{tabular}{l | cc ccc}
\toprule
\textbf{Method} & {Params(M)} & {Ops(G)} & {PSNR} & {SSIM} & {LPIPS} \\
\midrule
gcd/8  & 0.59 & 1.61 & 21.09 & 0.7873 & 0.2899 \\
gcd/4  & 0.44 & 1.14 & 20.98 & 0.7870 & 0.2906 \\
gcd/2  & 0.36 & 0.90 & 20.80 & 0.7867 & 0.2939 \\
gcd    & 0.31 & 0.78 & \textbf{20.97} & \textbf{0.7856} & \textbf{0.2845} \\
\bottomrule
\end{tabular}
}
\vspace{-4.mm}
\end{table*}

\begin{figure*}[t]
    \centering
    \begin{minipage}[t]{0.485\textwidth}
        \centering
        \includegraphics[width=\linewidth]{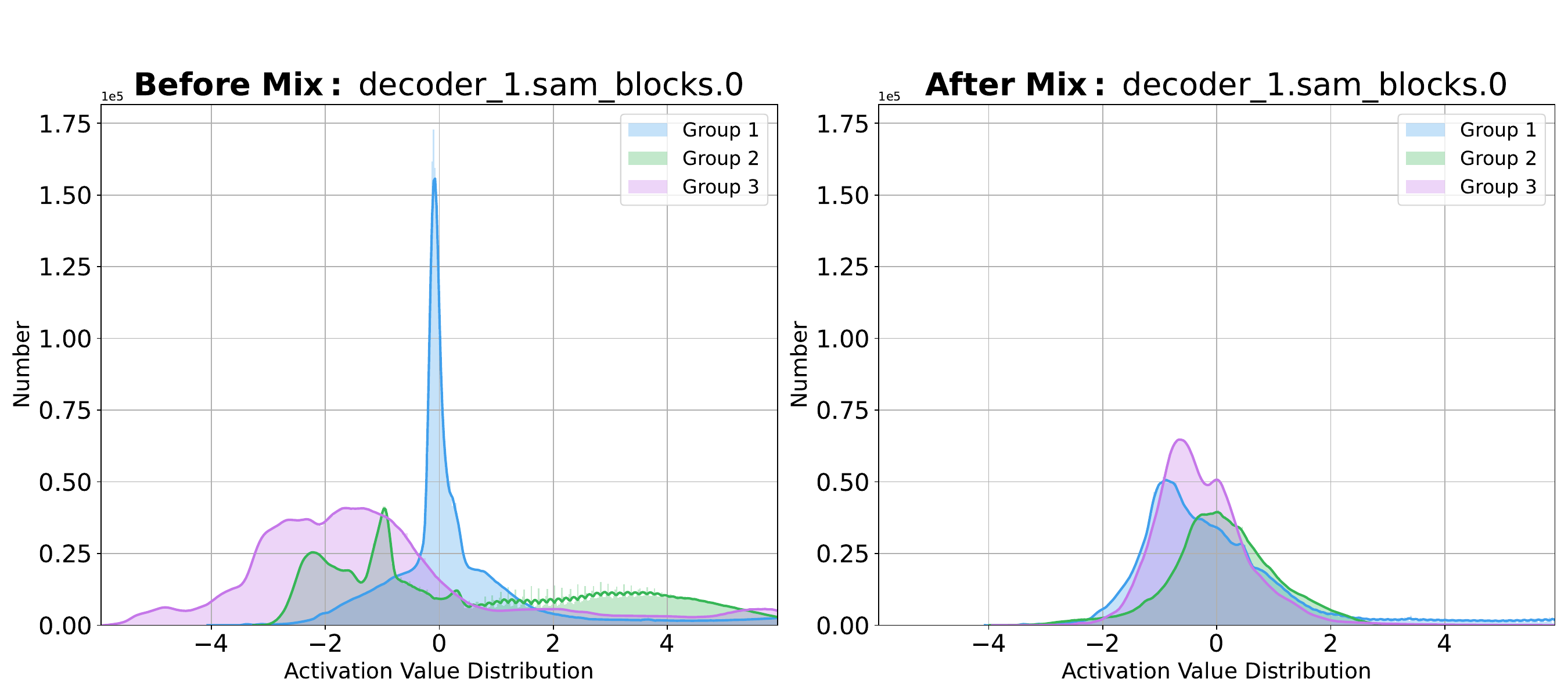}
    \end{minipage}
    \hfill
    \begin{minipage}[t]{0.485\textwidth}
        \centering
        \includegraphics[width=\linewidth]{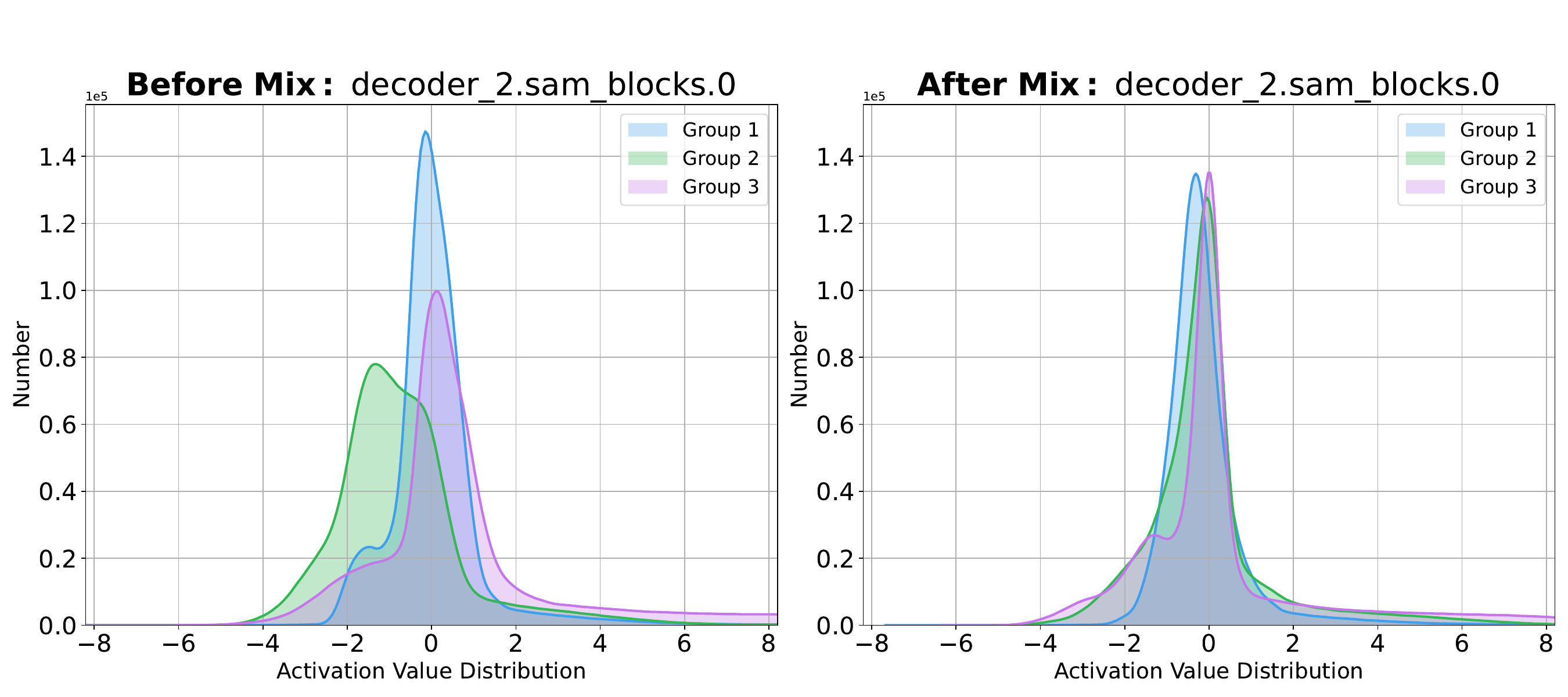}
    \end{minipage}
    \caption{Compare of activation distribution with (w/, right) and without (w/o, left) the interleaved mixing operation. For ease of illustration, we only present the activation distributions of the first three groups.}
    \label{fig:group}
    \vspace{-4.mm}
\end{figure*}

\begin{figure}[t]
    \vspace{1.mm}
    \centering
    \begin{subfigure}[t]{0.32\linewidth}
        \centering
        \includegraphics[width=\linewidth]{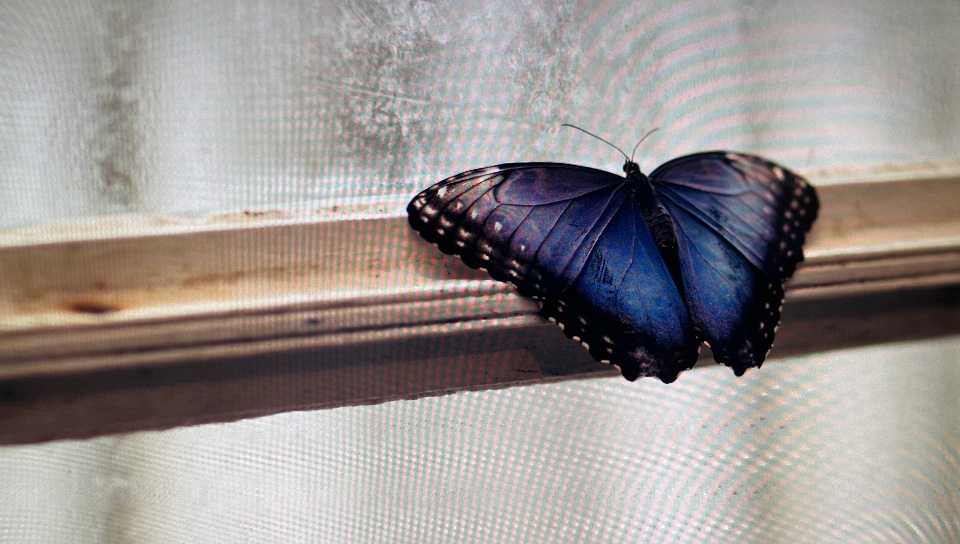}
        \caption{Input}
    \end{subfigure}\hfill
    \begin{subfigure}[t]{0.32\linewidth}
        \centering
        \includegraphics[width=\linewidth]{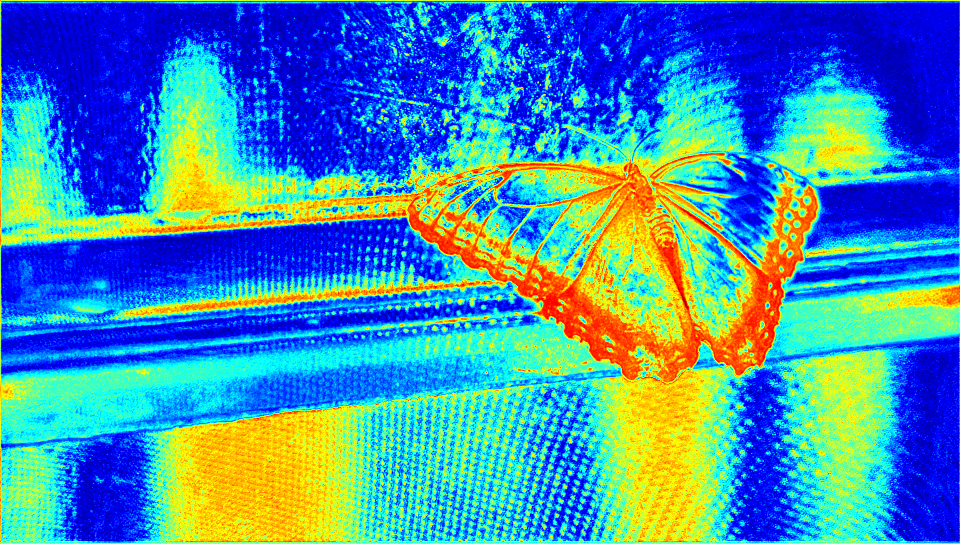}
        \caption{w/o MABG}
    \end{subfigure}\hfill
    \begin{subfigure}[t]{0.32\linewidth}
        \centering
        \includegraphics[width=\linewidth]{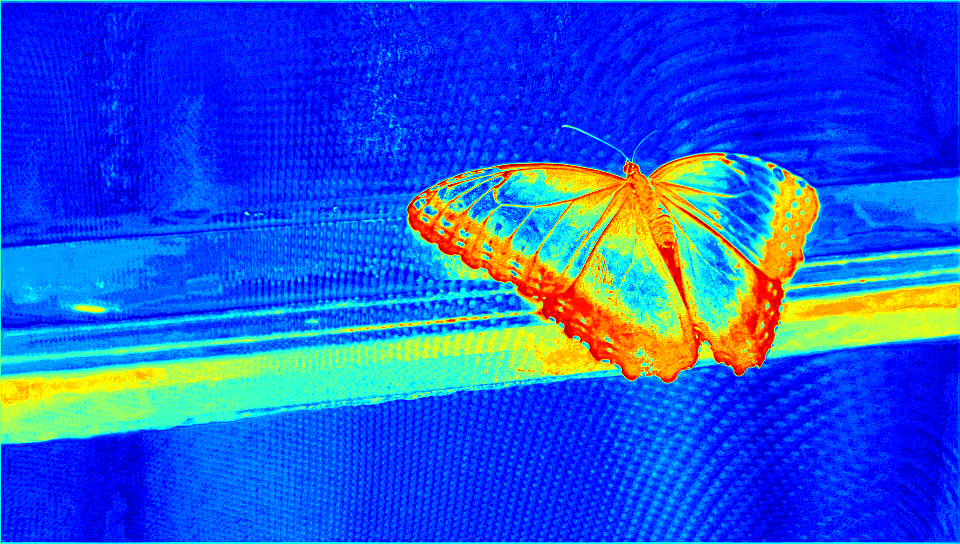}
        \caption{w/ MABG}
    \end{subfigure}

    \vspace{0.3mm}

    \begin{subfigure}[t]{0.32\linewidth}
        \centering
        \includegraphics[width=\linewidth]{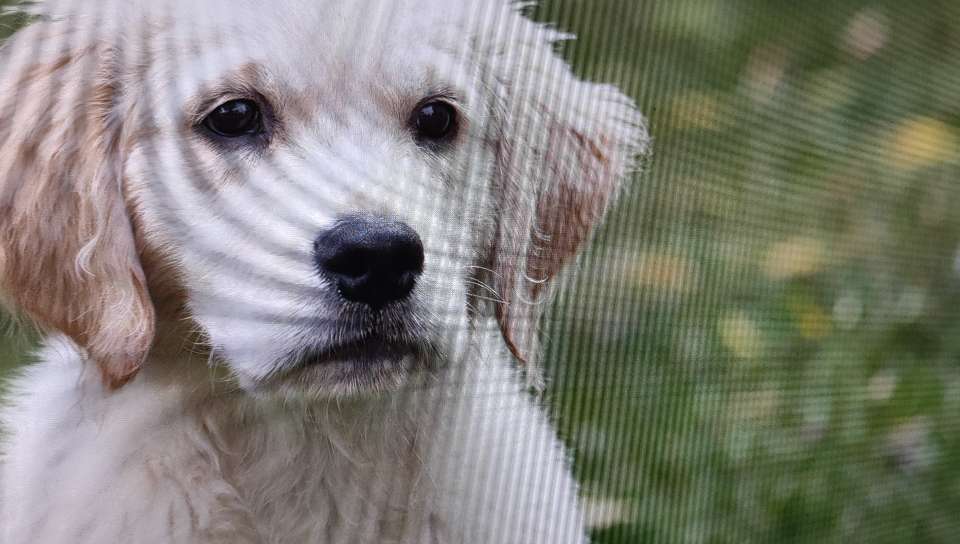}
        \caption{Input}
    \end{subfigure}\hfill
    \begin{subfigure}[t]{0.32\linewidth}
        \centering
        \includegraphics[width=\linewidth]{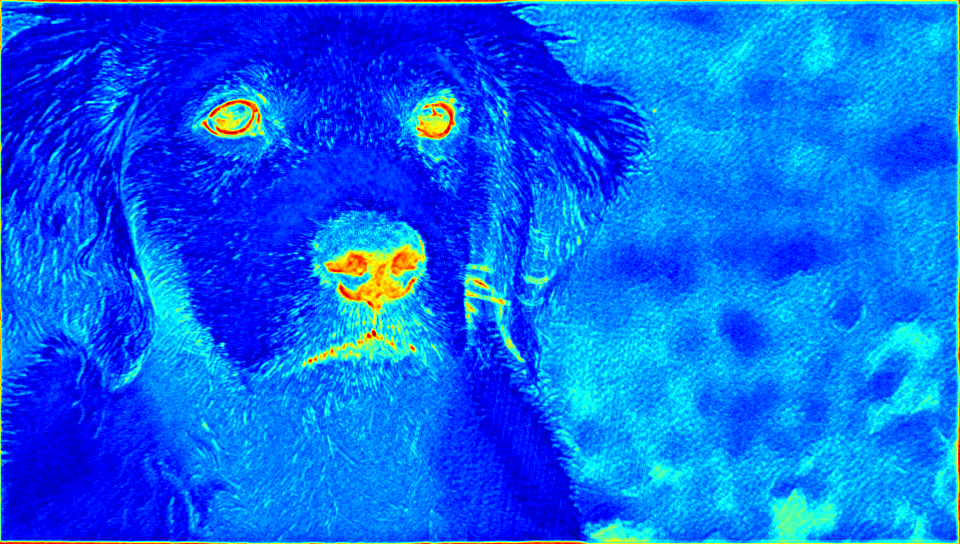}
        \caption{w/o MABG}
    \end{subfigure}\hfill
    \begin{subfigure}[t]{0.32\linewidth}
        \centering
        \includegraphics[width=\linewidth]{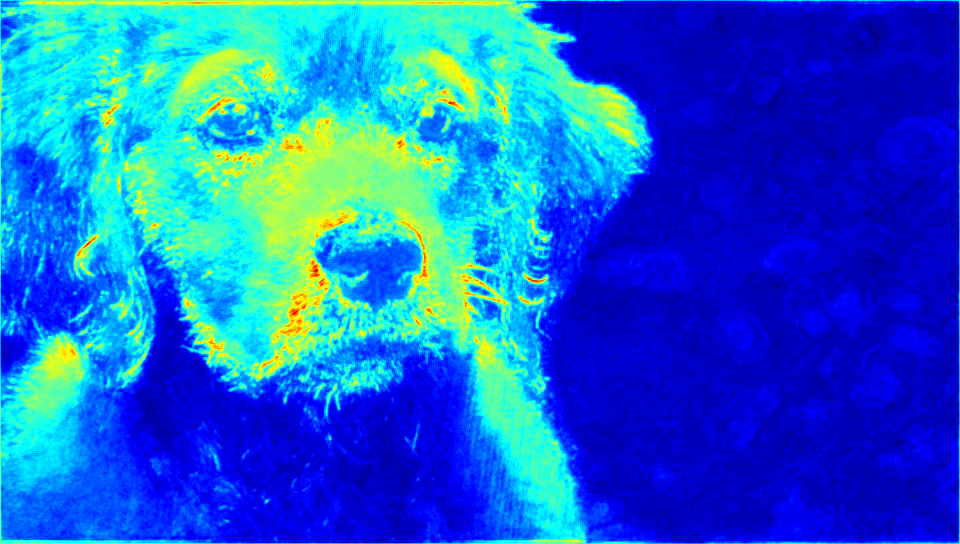}
        \caption{w/ MABG}
    \end{subfigure}

    \caption{Visualization of feature maps before and after applying MABG. Incorporating MABG enables the module to better focus on the main structure while reducing background interference.}
    \label{fig:feature_visual}
    \vspace{-6.mm}
\end{figure}

\vspace{-1.mm}
\subsection{Ablation Study}
\vspace{-1.mm}
We perform all ablation experiments on the UHDM dataset. 
Unless stated otherwise, we follow the identical training procedure as described in Sec.~\ref{sec:setting}, and train all variants for 150 epochs. 
We report the computational complexity in terms of operations (OPs) for a single forward pass, measured on an input of size $3\times256\times256$.

\textbf{Break Down.}
We first examine the effect of each proposed component in Tab.~\ref{tab:ablation_break}. The baseline is the binarized convolution block from BBCU~\cite{xia2022basic} (without MABG and SGRA). Introducing MABG yields a clear improvement in restoration quality, indicating that frequency-conditioned gating reduces critical errors for demoiréing. SGRA further strengthens performance by providing a lightweight, reliable shortcut alignment under frequent dimension transitions. Notably, combining MABG and SGRA achieves the best results, suggesting that modulation in the binarized main branch and structured shortcut adaptation complement.

As shown in Fig.~\ref{fig:feature_visual}, when MABG is applied, the feature maps exhibit a clearer representation of high-frequency components and sharper details, effectively reducing noise. In contrast, without MABG, the feature maps appear more blurred, with a loss of high-frequency information.

\textbf{Moir\'e-Aware Binary Gate (MABG).}
We conduct ablations on MABG in Tab.~\ref{tab:ablation_mabg} to verify the effectiveness of its conditioning design. Specifically, we evaluate three simplified variants: (i) conditioning the gate only on frequency descriptors, (ii) conditioning only on activation statistics, and (iii) replacing the conditioned gate with a purely learnable (input-agnostic) gate. The results show that using frequency cues alone or statistics alone consistently underperforms the full design that combines both, indicating that the two sources of information are complementary and jointly important for capturing structured demoir\'e degradations. Moreover, the purely learnable gate yields inferior performance, suggesting that unguided parameterization is insufficient in this setting. Overall, these comparisons validate that the proposed frequency-and-statistics conditioning is critical for effective gate modulation in binary convolution.

\textbf{Shuffle-Grouped Residual Adapter (SGRA).}
We analyze SGRA from three points of view. First, we ablate the proposed interleaved mixing by disabling channel reorganization (w/o Mix) and observe a clear degradation as shown in Tab.~\ref{tab:ablation_sgra}, indicating that explicit cross-partition interaction is important for preserving shortcut information beyond independent partition-wise projections. 

The visualization in Fig.~\ref{fig:group} further illustrates the effect of interleaved mixing by presenting the activation value distributions of the first three channel groups. Before applying Interleaved Mixing, different groups exhibit noticeable distribution shifts and distinct activation patterns, indicating limited interaction and feature sharing across groups. After channel interleaving is introduced, the activation distributions become more aligned and overlapping, suggesting that group discrepancies are effectively reduced.

Second, we compare SGRA with a low-rank shortcut projection where the rank is fixed to $2$ to keep the parameter budget comparable. Although both approaches achieve a similar level of compression, the low-rank variant yields inferior performance, as shown in Tab.~\ref{tab:ablation_sgra}. This result suggests that SGRA offers a more effective and structurally better-suited mechanism for accommodating frequent channel-width transitions in the binarized backbone.

Finally, we set the partition number to the greatest common divisor of the input and output channels. This choice allows an even split on both sides and leads to stable residual alignment in practice. In contrast, deviating from this setting (e.g., adopting smaller partitions) does not bring consistent improvements and may introduce unnecessary computational overhead, as shown in Tab.~\ref{tab:ablation_gcd}.

\subsection{Comparison with State-of-the-Art Methods} 
We evaluate our suggested BinaryDemoire against recent binarization methods, including ReActNet~\cite{liu2020reactnet}, BBCU~\cite{xia2022basic}, BiSRNet~\cite{cai2023binarized}, Biper~\cite{vargas2024biper}, and BiMaCoSR~\cite{liu2025bimacosr}. All binarization algorithms apply ESDNet~\cite{yu2022towards} as the base model. To guarantee a balanced comparison, we match the model size by adjusting the Params of all binarization methods to a similar level. 

\vspace{-0.5mm}
In addition we compare BinaryDemoire with some of full-precision demoir\'ing models, include MopNet~\cite{he2019mop}, MDDM~\cite{cheng2019multi}, FHDe$^2$Net~\cite{he2020fhde2net}, and our base model ESDNet~\cite{yu2022towards}. These comparisons further reflect the performance of our method.

\begin{table*}[t]
\centering
\scriptsize
\setlength{\tabcolsep}{6pt}
\renewcommand{\arraystretch}{1.15}

\caption{Comparison with five 1-bit methods and four 32-bit models on four demoir\'eing benchmarks. ESDNet is the full-precision base model. The highest and second-highest results among binarization methods are colored with \textcolor{red}{red} and \textcolor{blue}{blue}.}
\resizebox{\textwidth}{!}{%
\begin{tabular}{l | cc ccc ccc cc cc}
\toprule
\multirow{2}{*}{\centering\textbf{Method}} & \textbf{Params} & \textbf{Ops} &
\multicolumn{3}{c}{\textbf{UHDM}} &
\multicolumn{3}{c}{\textbf{FHDMi}} &
\multicolumn{2}{c}{\textbf{TIP2018}} &
\multicolumn{2}{c}{\textbf{LCDMoire}} \\
& \textbf{(M)} & \textbf{(G)} &
\textbf{PSNR} & \textbf{SSIM} & \textbf{LPIPS} &
\textbf{PSNR} & \textbf{SSIM} & \textbf{LPIPS} &
\textbf{PSNR} & \textbf{SSIM} &
\textbf{PSNR} & \textbf{SSIM} \\
\midrule

MopNet~\cite{he2019mop}        & 58.57& 197.85      & 19.49 & 0.7572 & 0.3857 & 22.76 & 0.7958 & 0.1794 & 27.75 & 0.8950 & -     & -\\

MDDM~\cite{cheng2019multi}          & 7.64 &  30.66     & 20.09 & 0.7441 & 0.3409 & 20.83 & 0.7343 & 0.2515 & -     & -      & 42.49   & 0.9940\\
FHDe$^2$Net~\cite{he2020fhde2net}     & 13.57& 360.61      & 20.34 & 0.7496 & 0.3519 & 22.93 & 0.7885 & 0.1688 & 27.78 & 0.8960 & 41.40 & - \\
ESDNet~\cite{yu2022towards}        & 5.93 & 17.59 & 22.12 & 0.7956 & 0.2551 & 24.50 & 0.8351 & 0.1354 & 30.11 & 0.9200 & 45.34 & 0.9966 \\

\midrule
ReActNet~\cite{liu2020reactnet}  & 0.31 & 0.86  & 19.42 & 0.7274 & 0.3353 & 20.96 & 0.7629 & 0.2373 & 23.96 & 0.8141 & 32.75 & 0.9766 \\
BBCU~\cite{xia2022basic}      & 0.31 & 0.86  & 19.36 & 0.7251 & 0.3428 & 20.34 & 0.7665 & 0.2390 & 24.13 & 0.8040 & 32.12 & 0.9731 \\
BiSRNet~\cite{cai2023binarized}   & 0.33 & 0.86  & 19.70 & 0.7278 & 0.3305 & 21.26 & 0.7737 & 0.2250 & 24.47 & 0.8147 & 33.83 & 0.9811 \\
Biper~\cite{vargas2024biper}     & 0.31 & 0.86  & 19.46 & 0.7319 & 0.3308 & 21.05 & 0.7764 & 0.2222 & 23.86 & 0.8069 & 32.64 & 0.9761 \\
BiMaCoSR~\cite{liu2025bimacosr}  & 0.31 & 0.85  &
\textcolor{blue}{20.50} & \textcolor{blue}{0.7782} & \textcolor{blue}{0.2896} &
\textcolor{blue}{22.05} & \textcolor{blue}{0.7943} & \textcolor{blue}{0.1933} &
\textcolor{blue}{26.04} & \textcolor{blue}{0.8551} &
\textcolor{blue}{37.70} & \textcolor{blue}{0.9900} \\
BinaryDemoire (ours) & 0.31 & 0.78  &
\textcolor{red}{20.97} & \textcolor{red}{0.7856} & \textcolor{red}{0.2845} &
\textcolor{red}{22.47} & \textcolor{red}{0.8091} & \textcolor{red}{0.1661} &
\textcolor{red}{27.33} & \textcolor{red}{0.8795} &
\textcolor{red}{39.88} & \textcolor{red}{0.9914} \\
\bottomrule
\end{tabular}
}

\label{tab:sota}
\end{table*}



\begin{figure*}[!t]
\scriptsize
\centering
\begin{tabular}{cccccccc}

\hspace{-0.48cm}
\begin{adjustbox}{valign=t}
\begin{tabular}{c}
\includegraphics[width=0.2143\textwidth]{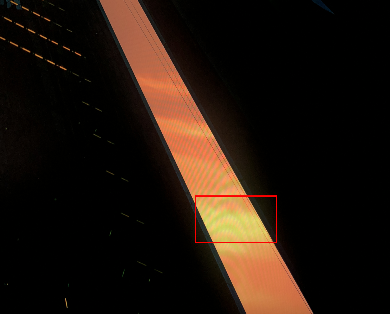}
\\
FHDMi: 01645
\end{tabular}
\end{adjustbox}
\hspace{-0.46cm}
\begin{adjustbox}{valign=t}
\begin{tabular}{cccccc}
\includegraphics[width=0.189\textwidth]{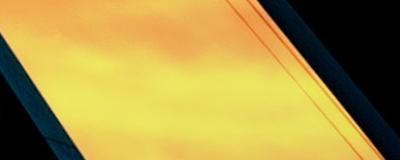} \hspace{-4.mm} &
\includegraphics[width=0.189\textwidth]{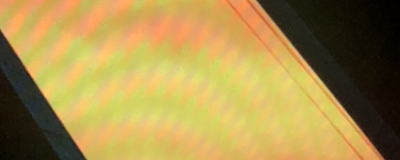} \hspace{-4.mm} &
\includegraphics[width=0.189\textwidth]{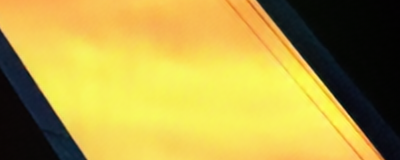} \hspace{-4.mm} &
\includegraphics[width=0.189\textwidth]{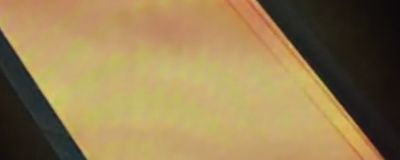} \hspace{-4.mm} &
\\ 
GT \hspace{-4.mm} &
INPUT \hspace{-4.mm} &
ESDNet (FP)~\cite{yu2022towards} \hspace{-4.mm} &
BBCU~\cite{xia2022basic} \hspace{-4.mm} &
\\
\includegraphics[width=0.189\textwidth]{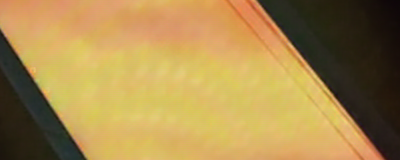} \hspace{-4.mm} &
\includegraphics[width=0.189\textwidth]{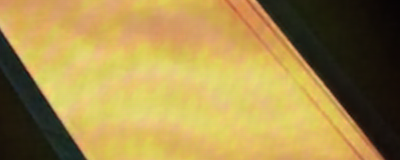} \hspace{-4.mm} &
\includegraphics[width=0.189\textwidth]{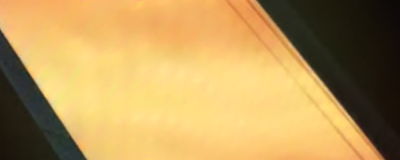} \hspace{-4.mm} &
\includegraphics[width=0.189\textwidth]{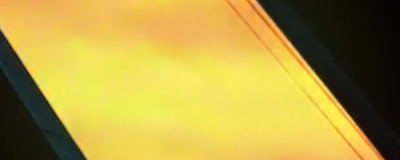} \hspace{-4.mm} &
\\ 
BiSRNet~\cite{cai2023binarized} \hspace{-4.mm} &
Biper~\cite{vargas2024biper} \hspace{-4.mm} &
BiMaCoSR~\cite{liu2025bimacosr} \hspace{-4.mm} &
BinaryDemoire (ours) \hspace{-4mm}
\\
\end{tabular}
\end{adjustbox}
\\

\hspace{-0.48cm}
\begin{adjustbox}{valign=t}
\begin{tabular}{c}
\includegraphics[width=0.2143\textwidth]{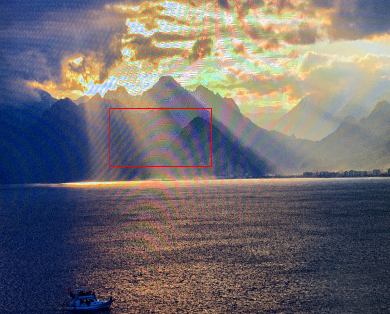}
\\
UHDM: 0495
\end{tabular}
\end{adjustbox}
\hspace{-0.46cm}
\begin{adjustbox}{valign=t}
\begin{tabular}{cccccc}
\includegraphics[width=0.189\textwidth]{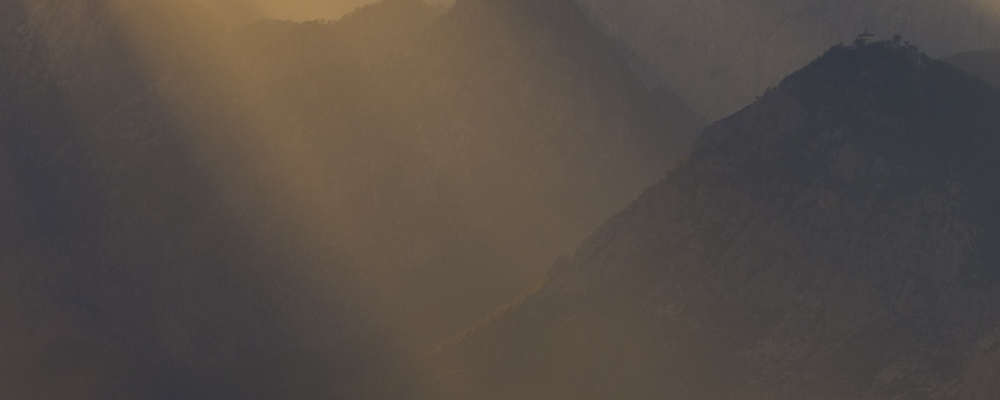} \hspace{-4.mm} &
\includegraphics[width=0.189\textwidth]{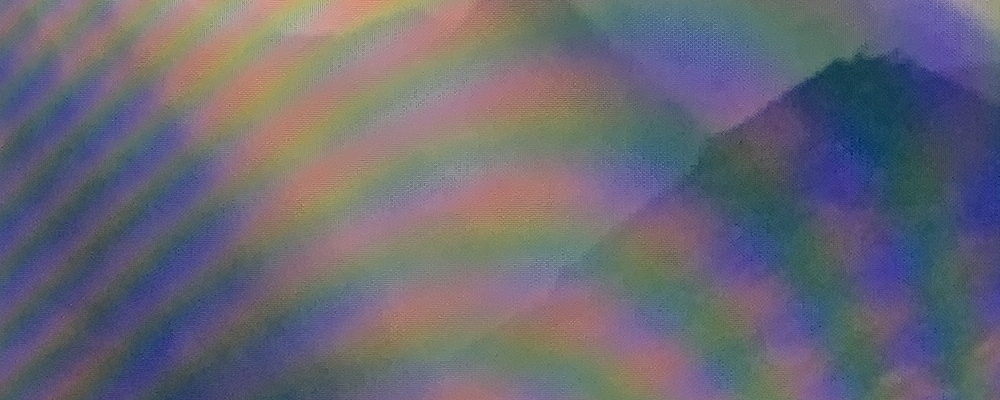} \hspace{-4.mm} &
\includegraphics[width=0.189\textwidth]{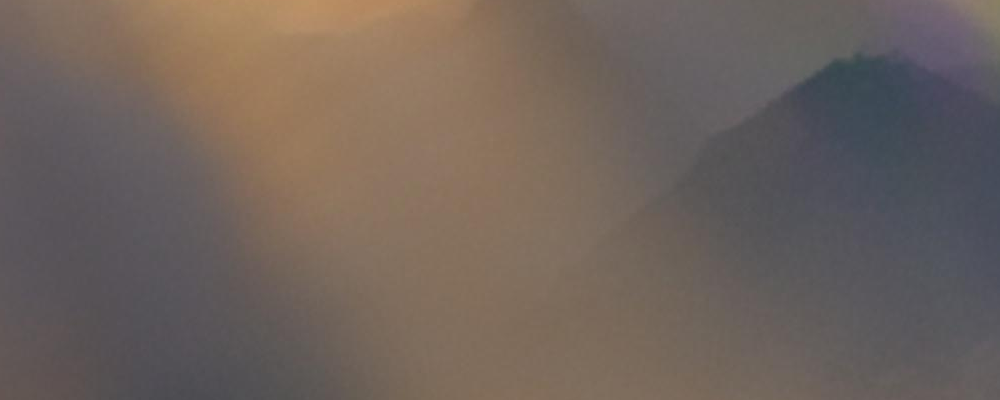} \hspace{-4.mm} &
\includegraphics[width=0.189\textwidth]{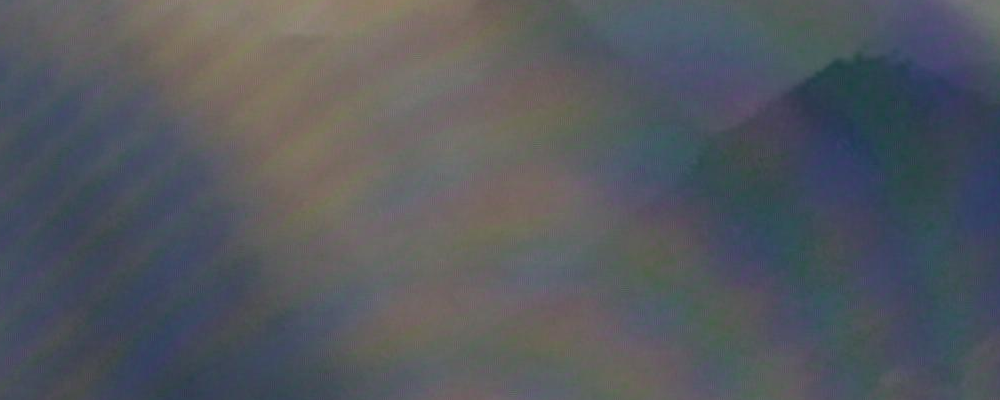} \hspace{-4.mm} &
\\ 
GT \hspace{-4.mm} &
INPUT \hspace{-4.mm} &
ESDNet (FP)~\cite{yu2022towards} \hspace{-4.mm} &
BBCU~\cite{xia2022basic} \hspace{-4.mm} &
\\
\includegraphics[width=0.189\textwidth]{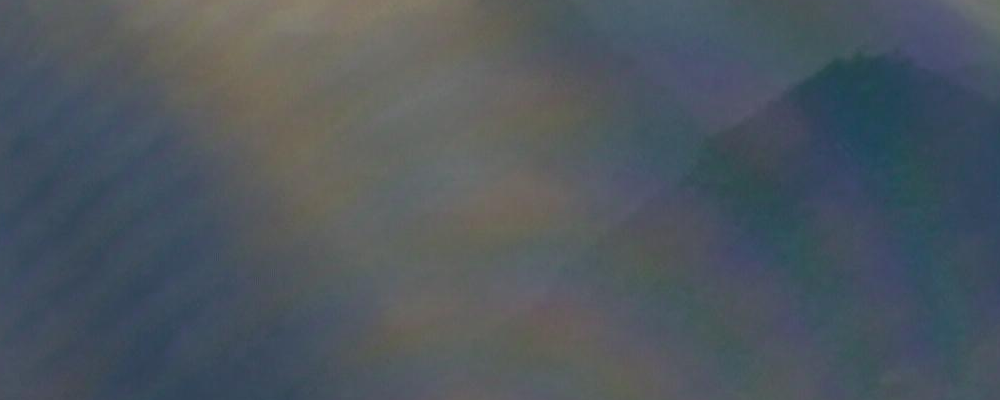} \hspace{-4.mm} &
\includegraphics[width=0.189\textwidth]{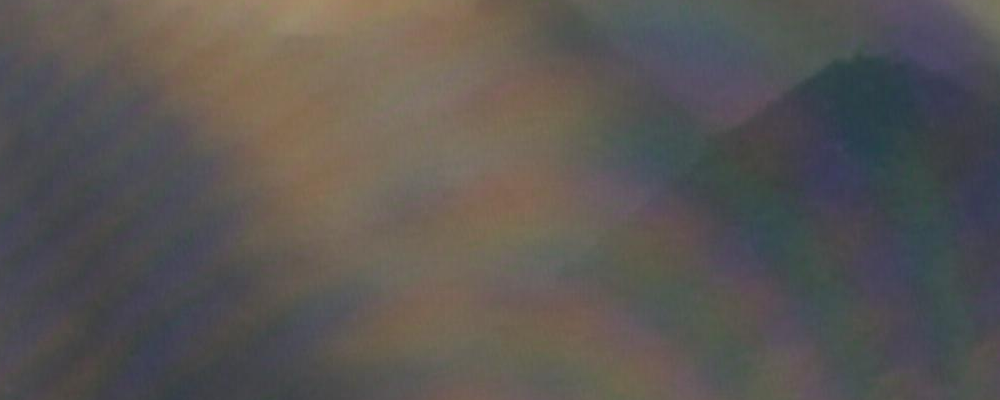} \hspace{-4.mm} &
\includegraphics[width=0.189\textwidth]{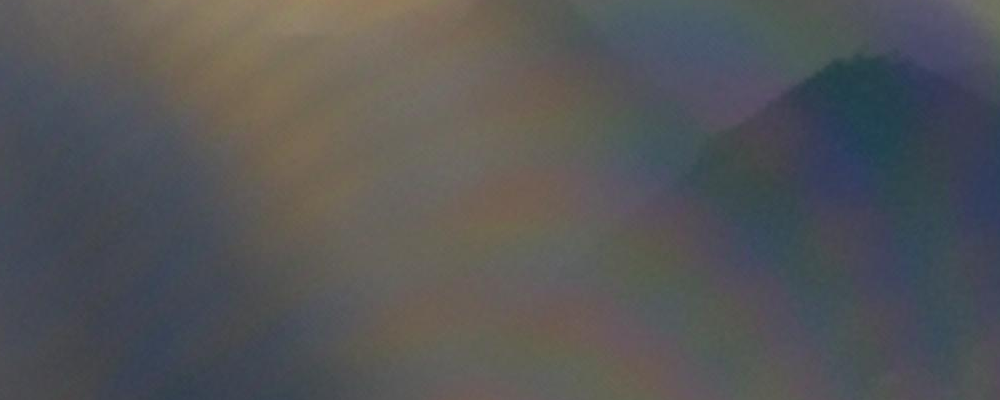} \hspace{-4.mm} &
\includegraphics[width=0.189\textwidth]{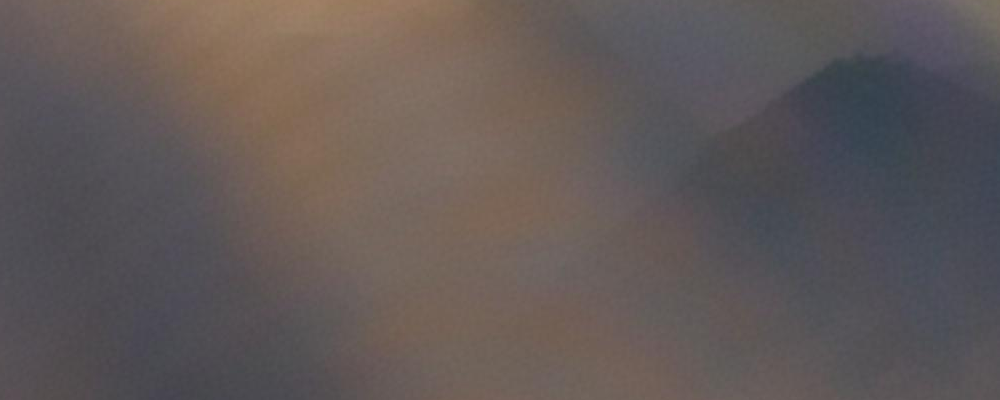} \hspace{-4.mm} &
\\ 
BiSRNet~\cite{cai2023binarized} \hspace{-4.mm} &
Biper~\cite{vargas2024biper} \hspace{-4.mm} &
BiMaCoSR~\cite{liu2025bimacosr} \hspace{-4.mm} &
BinaryDemoire (ours) \hspace{-4mm}
\\
\end{tabular}
\end{adjustbox}
\\

\hspace{-0.48cm}
\begin{adjustbox}{valign=t}
\begin{tabular}{c}
\includegraphics[width=0.2143\textwidth]{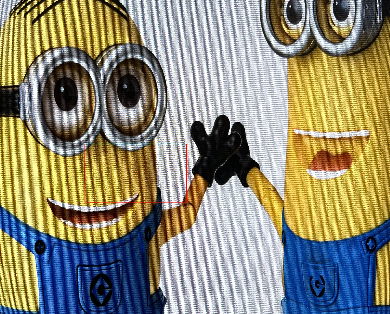}
\\
UHDM: 0124
\end{tabular}
\end{adjustbox}
\hspace{-0.46cm}
\begin{adjustbox}{valign=t}
\begin{tabular}{cccccc}
\includegraphics[width=0.189\textwidth]{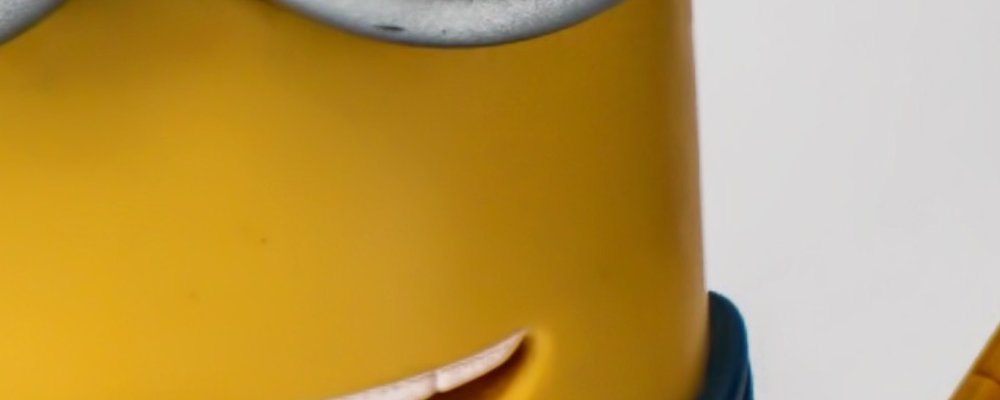} \hspace{-4.mm} &
\includegraphics[width=0.189\textwidth]{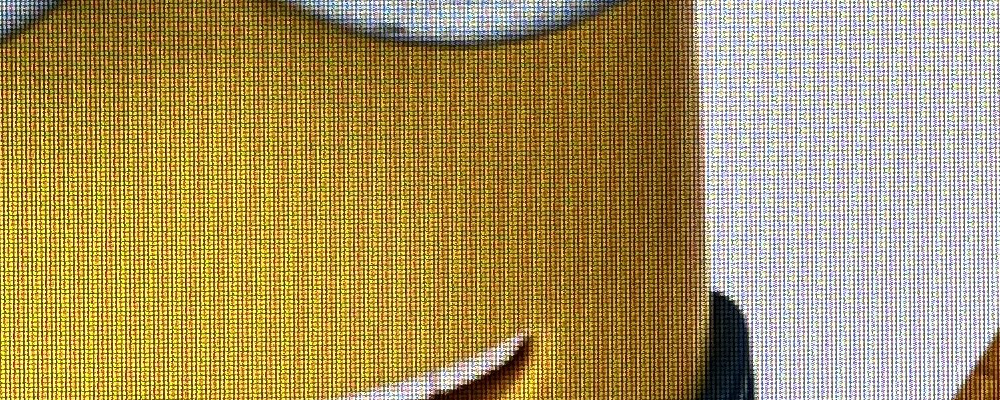} \hspace{-4.mm} &
\includegraphics[width=0.189\textwidth]{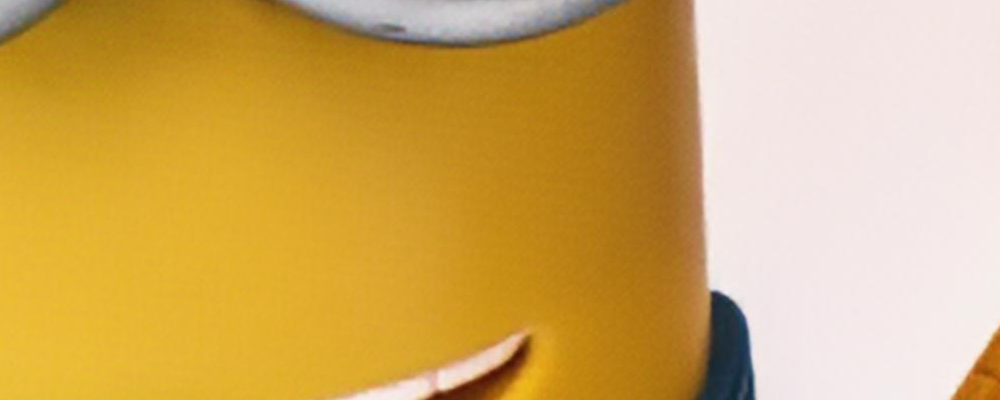} \hspace{-4.mm} &
\includegraphics[width=0.189\textwidth]{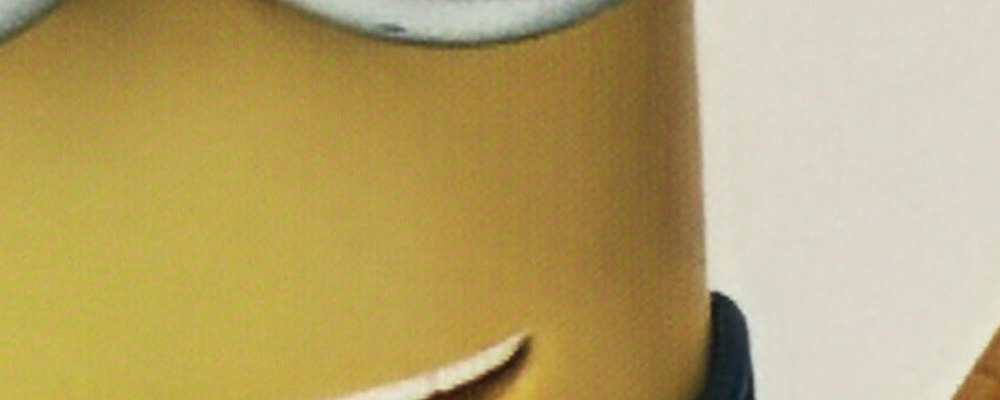} \hspace{-4.mm} &
\\ 
GT \hspace{-4.mm} &
INPUT \hspace{-4.mm} &
ESDNet (FP)~\cite{yu2022towards} \hspace{-4.mm} &
BBCU~\cite{xia2022basic} \hspace{-4.mm} &
\\
\includegraphics[width=0.189\textwidth]{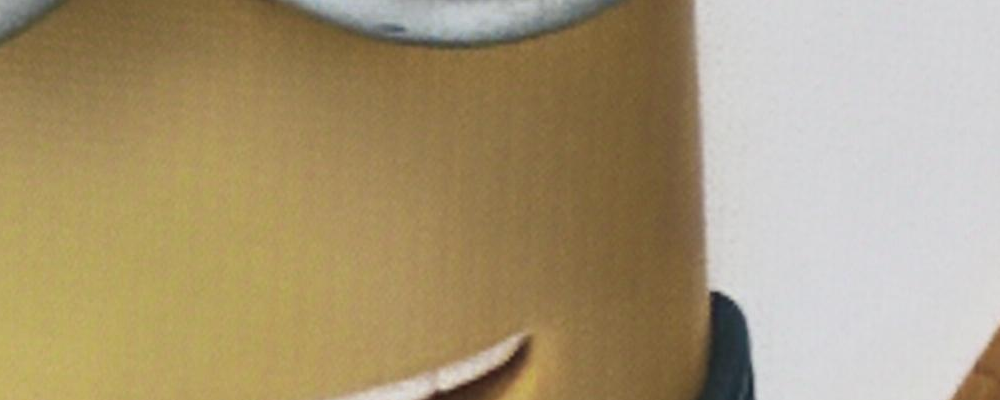} \hspace{-4.mm} &
\includegraphics[width=0.189\textwidth]{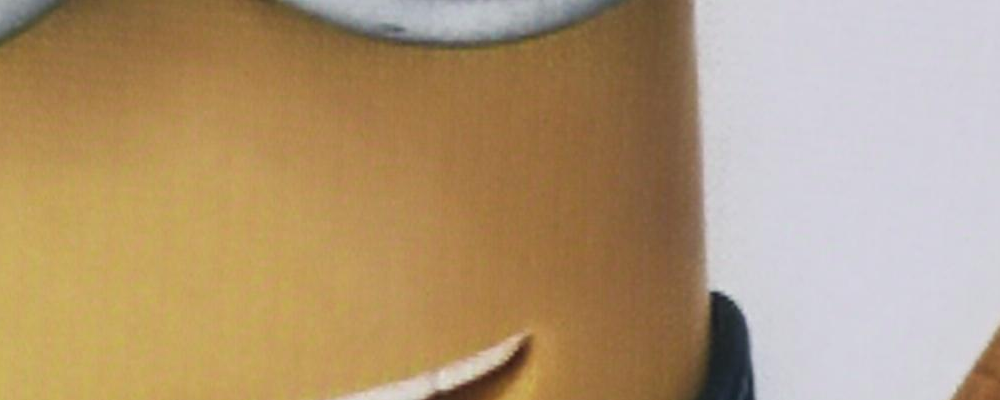} \hspace{-4.mm} &
\includegraphics[width=0.189\textwidth]{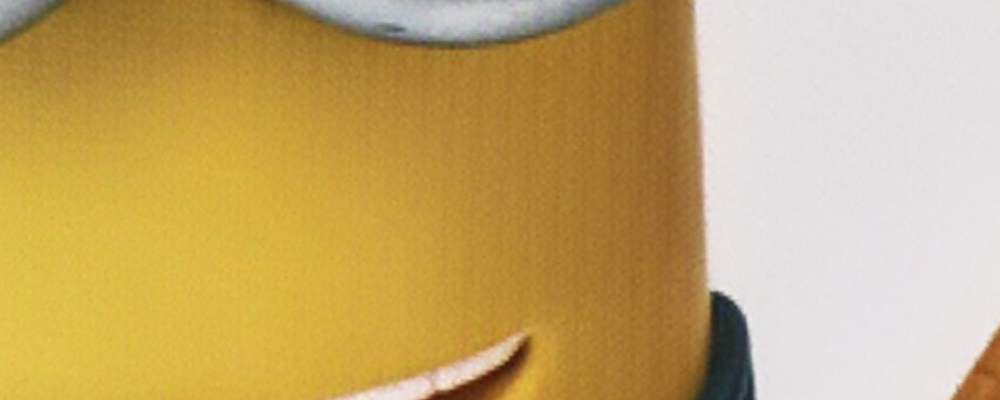} \hspace{-4.mm} &
\includegraphics[width=0.189\textwidth]{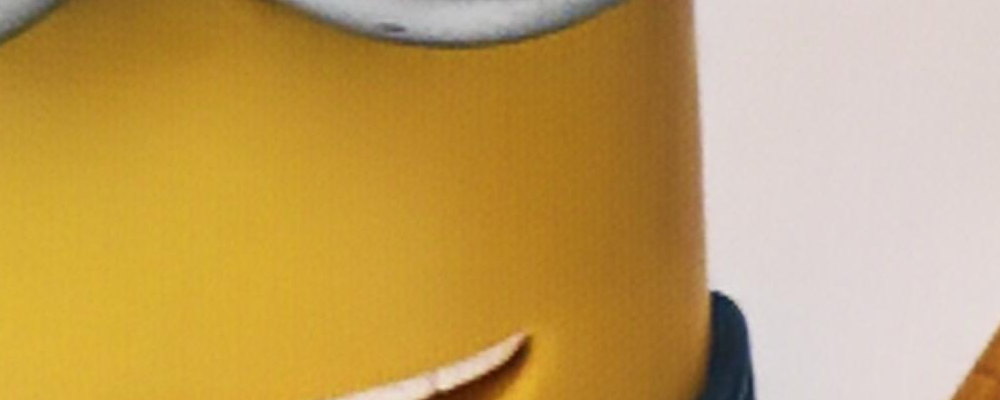} \hspace{-4.mm} &
\\ 
BiSRNet~\cite{cai2023binarized} \hspace{-4.mm} &
Biper~\cite{vargas2024biper} \hspace{-4.mm} &
BiMaCoSR~\cite{liu2025bimacosr} \hspace{-4.mm} &
BinaryDemoire (ours) \hspace{-4mm}
\\
\end{tabular}
\end{adjustbox}
\\

\end{tabular}
\caption{Visual comparison on different datasets, our BinaryDemoire outperforms recent binarization methods}
\label{fig:visual}
\vspace{-2.mm}
\end{figure*}

\vspace{-.5mm}
\textbf{Quantitative Results.}
We report quantitative comparisons in Tab.~\ref{tab:sota}. We calculate the OPs for a single forward pass with an output image size of $3\times256\times256$. Compared with the second-best binarized baseline (BiMaCoSR), our BinaryDemoire achieves higher PSNR on both UHDM and TIP2018, with gains of \textbf{+0.47 dB} and \textbf{+1.29 dB}. Moreover, BinaryDemoire demonstrates competitive performance against existing full-precision methods. For instance, compared with FHDe$^2$Net~\cite{he2020fhde2net}, our method outperforms it on most evaluation metrics on FHDMi.

\vspace{-1.mm}
In terms of efficiency, BinaryDemoire uses $0.31$M parameters and $0.78$G OPs, which correspond to only \textbf{5.2\%} of the full-precision base model parameters (a \textbf{94.8\%} reduction) and \textbf{4.4\%} of its OPs (a \textbf{95.6\%} reduction). These impressive results indicate that our BinaryDemoire achieves remarkable performance-efficiency balance under extreme compression, demonstrating its effectiveness and practicality.

\textbf{Visual Results.}
Figure.~\ref{fig:visual} presents visual comparisons of our BinaryDemoire with existing methods across multiple datasets. As shown, our approach is able to restore finer textures and structural details while effectively suppressing complex moir\'e patterns. In contrast, existing binarized methods tend to introduce artifacts or color distortions. For example, in the first case, compared methods fail to remove severe moiré patterns, whereas our method successfully eliminates the artifacts while preserving accurate color.

Meanwhile, BinaryDemoire achieves outcomes that are much closer to the full-precision version, demonstrating its effectiveness in restoration and computational efficiency. More results are included in the supplementary material.

\section{Conclusion}
\vspace{-1.mm}
In this paper, we propose the BinaryDemoire, a novel binarized model for image demoir\'eing. To address the issue of binarization being sensitive to high-frequency noise,  we propose the moir\'e-aware binary gate (MABG). The module adaptively modulates binary convolution responses using lightweight frequency descriptors and activation statistics. We further designed a shuffle-grouped residual adapter (SGRA) to efficiently align residual connections under frequent dimensional mismatches. Extensive experiments on multiple benchmarks demonstrate that BinaryDemoire outperforms existing binarization methods, achieving superior restoration quality with high efficiency.


\bibliography{example_paper}
\bibliographystyle{icml2026}

\end{document}